\documentclass[11pt]{article}

\usepackage{acl}

\usepackage{times}
\usepackage{latexsym}

\usepackage[T1]{fontenc}
\usepackage[utf8]{inputenc}

\usepackage{microtype}

\usepackage{inconsolata}

\usepackage{graphicx}

\usepackage{hyperref}
\usepackage{url}
\usepackage{amssymb}
\usepackage{algorithm}
\usepackage{algpseudocode}
\usepackage{booktabs} 
\usepackage{multirow} 
\usepackage{siunitx}  
\usepackage{wrapfig}   
\usepackage{enumitem}
\usepackage{tcolorbox}

\usepackage{amsmath}
\usepackage{amssymb}
\usepackage{mathtools}
\usepackage{amsthm}

\usepackage{subcaption}

\newtheorem{theorem}{Theorem}

\renewcommand{\eqref}[1]{(\ref{#1})}

\usepackage{tcolorbox}
\tcbuselibrary{skins,breakable}

\usepackage[utf8]{inputenc}
\usepackage{xcolor} 
\usepackage{enumitem}
\newtcolorbox{examplebox}[1]{
  enhanced,
  colback=white,
  colframe=blue!80!black,
  boxrule=3pt,
  fonttitle=\bfseries,
  colbacktitle=blue!80!black,
  coltitle=white,
  sharp corners,
  title={#1},
  boxed title style={sharp corners,boxrule=1pt},
  breakable
}

\newcommand*\circled[1]{\tikz[baseline=(char.base)]{\node[shape=circle,draw,inner sep=0.2pt] (char) {#1};}}

\newtcolorbox{casebox}[1]{%
  enhanced,
  breakable,
  colback=white,
  colframe=blue!80!black,
  boxrule=1.2pt,
  fonttitle=\bfseries,
  colbacktitle=blue!80!black,
  coltitle=white,
  sharp corners,
  title={#1},
  left=1em,right=1em,top=0.8em,bottom=0.8em,
  before skip=6pt, after skip=6pt,
}

\newcommand{\hlblock}[2][yellow!25]{%
  \par\noindent
  \begingroup
  \setlength{\fboxsep}{1pt}%
  \colorbox{#1}{%
    \parbox{\dimexpr\linewidth-2\fboxsep\relax}{#2}%
  }%
  \endgroup
  \par
}
\title{Self-Reflective Generation at Test Time}

\author{
    \textbf{Jian Mu\textsuperscript{1}},
    \textbf{Qixin Zhang\textsuperscript{2}},
    \textbf{Zhiyong Wang\textsuperscript{3}}, 
    \textbf{Menglin Yang\textsuperscript{1}}, 
    \textbf{Shuang Qiu\textsuperscript{4}},
    \\
    \textbf{Chengwei Qin\textsuperscript{1}}, 
    \textbf{Zhongxiang Dai\textsuperscript{5}}, 
    \textbf{Yao Shu\textsuperscript{1}}\thanks{Corresponding author}
\\
\textsuperscript{1}Hong Kong University of Science and Technology (Guangzhou), \\
\textsuperscript{2}Nanyang Technological University, \textsuperscript{3}University of Edinburgh,\\
\textsuperscript{4}City University of Hong Kong, \textsuperscript{5}The Chinese University of Hong Kong, Shenzhen \\
\texttt{jianmu@hkust-gz.edu.cn}, \texttt{yaoshu@hkust-gz.edu.cn}
}
\usepackage{booktabs}
\usepackage{multirow}
\usepackage{siunitx}
\usepackage[table]{xcolor}
\definecolor{srgengray}{gray}{0.92} % SRGen 行的背景色
\definecolor{deltagray}{gray}{0.5}  % 涨幅数字的颜色

\newcommand{\best}[1]{\textbf{#1}} 
\newcommand{\deltaT}[1]{\textcolor{deltagray}{\scriptsize~(#1)}}
\newcommand{\name}{SRGen}

\begin{document}
\maketitle
\begin{abstract}
Large language models (LLMs) increasingly solve complex reasoning tasks via long chain-of-thought, but their forward-only autoregressive generation process is fragile; early token errors can cascade, which creates a clear need for self-reflection mechanisms. However, existing self-reflection either performs revisions over full drafts or learns self-correction via expensive training, both fundamentally reactive and inefficient. To address this, we propose \textit{Self-Reflective Generation at Test Time} (\name{}), a lightweight test-time framework that reflects before generating at uncertain points. During token generation, SRGen utilizes dynamic entropy thresholding to identify high-uncertainty tokens. For each identified token, it trains a specific corrective vector, which fully exploits the already generated context for a self-reflective generation to correct the token probability distribution. By retrospectively analyzing the partial output, this self-reflection enables more trustworthy decisions, thereby significantly reducing the probability of errors at highly uncertain points. Evaluated on challenging mathematical reasoning benchmarks and a diverse set of LLMs, \name{} can significantly strengthen model reasoning.
Moreover, our findings position \name{} as a plug-and-play method that integrates reflection into the generation process for reliable LLM reasoning, achieving consistent gains with bounded overhead and can be combined with other training-time (e.g., RLHF) and test-time (e.g., SLOT) techniques.
\end{abstract}

\begingroup
\renewcommand{\thefootnote}{\fnsymbol{footnote}}
\footnotetext[2]{\url{https://github.com/2020-qqtcg/SRGen}}
\endgroup

\section{Introduction}

The ability to execute complex multi-step reasoning remains a central frontier in advancing large language models (LLMs). LLMs generate step-by-step reasoning traces, often called chain-of-thought (CoT)~\citep{wei2022chain}. This capability has enabled substantial progress in mathematics, program synthesis, and other domains~\citep{yao2023tree, plaat2024reasoning}. The fidelity of these traces often determines whether the final answer is correct~\citep{paul2024making, hammoud2025beyond}. Thus, improving the reliability of the reasoning process is critical to realizing the full potential of LLMs.

A fundamental tension persists between the fluid, self-corrective character of human problem solving and the rigid, forward-only dynamics of standard LLM decoding. Humans iterate: they pause, re-evaluate premises, and change course. In contrast, LLMs perform autoregressive decoding~\citep{vaswani2017attention}: each token depends on all preceding tokens, and prior outputs cannot be revised. As a result, early errors can propagate and compound, derailing the entire trajectory~\citep{jain-etal-2025-first}. This brittleness in forward-only decoding is a major obstacle to reliable reasoning.

Many prior works tackle this fragility via error correction. One line pursues post hoc iterative refinement: the model critiques and revises a complete draft in subsequent passes~\citep{madaan2023self, yuksekgonul2024textgrad}, incurring substantial latency and computational cost. Another line trains models for intrinsic self-correction, for example via reinforcement learning~\citep{bensal2025reflect, ma2025s}. This enables mid-reasoning fixes but still requires that an erroneous segment be produced before intervention. Crucially, both approaches are reactive; they address errors only after they have occurred. The challenge of proactive error prevention, namely steering the model away from a mistake before it is committed, remains a research gap.

To this end, we introduce \textit{\textbf{S}elf-\textbf{R}eflective \textbf{Gen}eration at Test Time} (\name{}), a lightweight inference-time framework for proactive error prevention. The key premise is that tokens differ in informativeness: recent work identifies ``\textit{critical tokens}'' via high predictive entropy~\citep{wang2025beyond}, low confidence~\citep{fu2025deep}, or spikes in mutual information~\citep{qian2025demystifying}. Instead of using these signals solely to adjust sampling or apply post hoc filtering, \name{} intervenes at the moment of risk: during generation it detects critical points, briefly pauses, and optimizes a small corrective vector $\delta$ with a token-level reflection loss; this vector is injected into the hidden state before emitting the next token. The intervention is local and transient, steering the model away from early errors without additional full passes.

\begin{table}[t]
\centering
\footnotesize 
\renewcommand{\arraystretch}{1.3}
\setlength{\tabcolsep}{4pt} 

\caption{Conceptual comparison of self-reflection paradigms. }
\label{tab:paradigm_comparison}

\begin{tabular}{
  c       % Paradigm
  c       % Timing
  c       % Train
  c       % Infer
  c       % Mode
}
\toprule
& & \multicolumn{2}{c}{\textbf{Cost}} & \\   % 第一行只写 Cost
\cmidrule(lr){3-4}
\textbf{Paradigm} & \textbf{Timing} & \textbf{Train} & \textbf{Infer} & \textbf{Mode} \\  % 第二行所有标题
\midrule

RL Correction 
& Training 
& High 
& Low 
& Reactive \\

Post-hoc Refine 
& Post-gen 
& Zero 
& High 
& Reactive \\

\rowcolor{srgengray}
\textbf{SRGen (Ours)} 
& \textbf{In-gen} 
& \textbf{Zero} 
& \textbf{Low} 
& \textbf{Proactive} \\

\bottomrule
\end{tabular}
\end{table}

Table~\ref{tab:paradigm_comparison} positions \name{} relative to existing self-reflection approaches and highlights a new paradigm of proactive, test-time self-reflection. \name{} requires no additional training (unlike RL-based methods), avoids the latency of post hoc iterative refinement, and prevents errors before they compound. Its plug-and-play nature makes it broadly applicable to pre-trained language models and compatible with other reasoning-enhancement techniques, including SFT, RL, and distillation trained models and with some test-time methods even using similar mechanisms.

Our main contributions are as follows:
\begin{itemize}[leftmargin=*,noitemsep,topsep=2pt]
  \item We propose SRGen, which consists of two core components: an uncertainty-based token monitor and a self-reflective corrective vector computed online. Without any external feedback, SRGen improves reasoning by reducing decision errors at uncertain tokens during generation.
  \item We provide theoretical motivation for our design and conduct analyses to support and interpret the proposed method.
  \item Extensive experiments demonstrate that SRGen achieves larger gains over strong baselines while introducing lower latency overhead. SRGen also composes well with related methods, yielding further improvements when combined.

\end{itemize}

\section{Problem Formulation}
In autoregressive language models, the generation of a token sequence \( y = (y_0, y_1, \dots, y_t) \) is modeled by the product of conditional probabilities:
\begin{equation}
P(y | x_0) = \prod_{t=1}^{T} P(y_t | y_{<t}, x_0; \theta),
\end{equation}
where \( x_0 \) is the input prompt and \( \theta \) denotes the model parameters. Such a forward-only decoding process, unfortunately, exhibits a key vulnerability: fragility. An early error in a reasoning chain can propagate and amplify, leading to catastrophic failures in the final output \citep{jain-etal-2025-first}.

Existing solutions predominantly fall into two categories.
\textbf{(1) Post-hoc Iterative Refinement.} Methods such as Self-Refine \citep{madaan2023self} employ a multi-stage pipeline where the model critiques and revises a complete draft \citep{shinn2023reflexion, yu2024teaching, yuksekgonul2024textgrad}. Although often effective, this approach incurs substantial computational overhead and latency.
\textbf{(2) Training for Intrinsic Self-Correction.} This line of work embeds correction capabilities directly into the model parameters, typically via techniques like reinforcement learning \citep{bai2022constitutional, hu2024teaching, kumar2024training, moskvoretskii2025self, bensal2025reflect}. These methods require expensive, resource-intensive training and can only intervene after an error has already been generated.

A common thread unites these approaches: their reactive nature, as they correct errors only after they have occurred. This limitation motivates our central research question:
\begin{tcolorbox}[colback=orange!5!white,colframe=orange!70!white, left=0.5mm, right=1mm, top=1mm, bottom=1mm]\label{insight-1}
{\fontfamily{ppl}\selectfont
\textit{Can we design a proactive error prevention mechanism that identifies and intervenes at potential error points in real-time during generation, thereby enhancing reasoning reliability within a single decoding pass and at a minimal additional cost?}}
\end{tcolorbox}

\section{Self-Reflection Generation Process}

\begin{figure*}[t]
\begin{center}
  \includegraphics[width=1\textwidth]{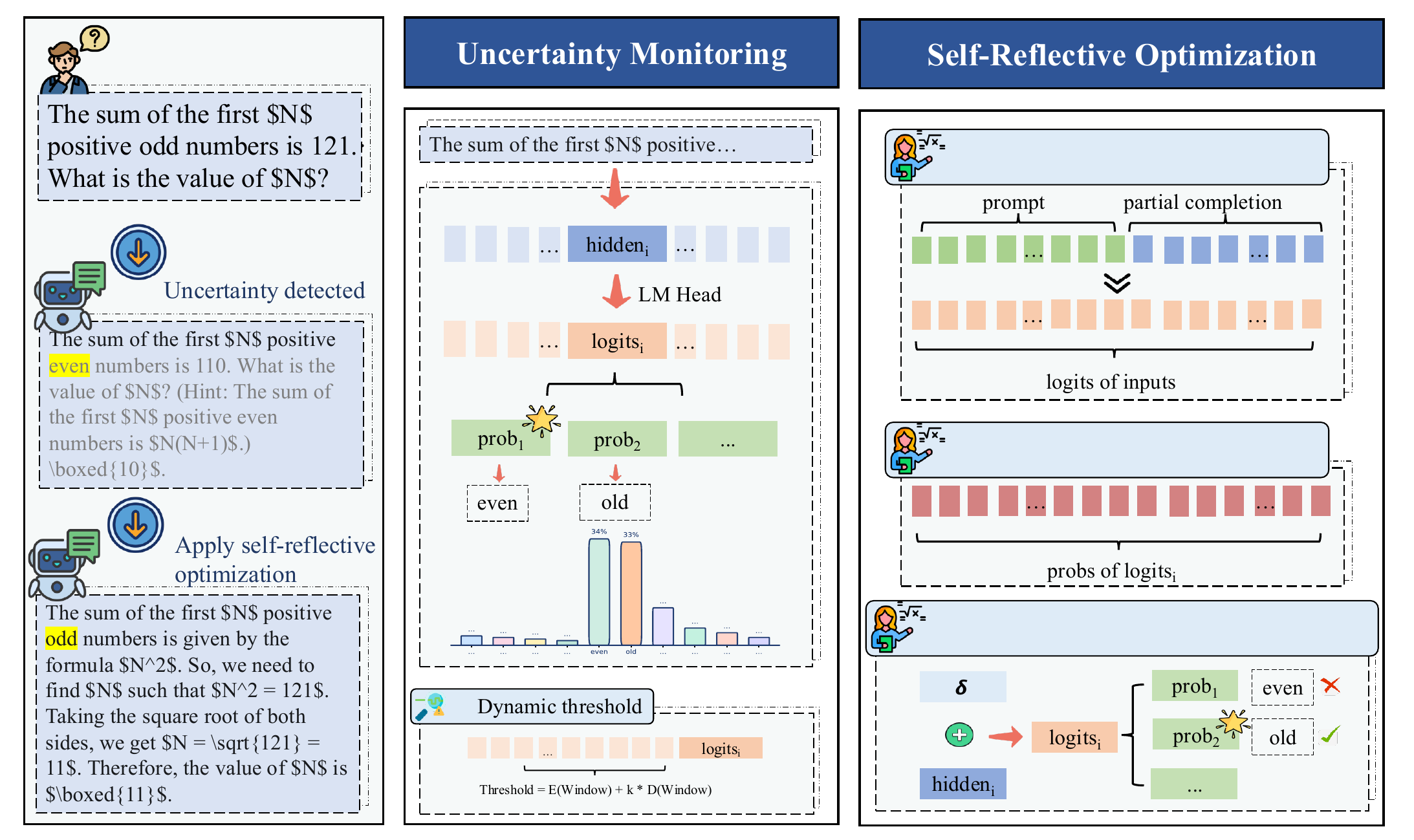}
\end{center}
\caption{An overview of the \textit{Self-Reflective Generation} (\name{}) framework. This framework consists of two main stages. (1) \textit{Uncertainty Monitoring}. A threshold is dynamically computed from the mean and standard deviation of token entropies within a recent history window of size N. (2) \textit{Self-Reflective Optimization}. If the current token's entropy exceeds the threshold, a correction vector, $\delta$, is optimized on-the-fly using a joint loss of cross-entropy and entropy minimization. This $\delta$ is then added to the token's hidden state to steer the final decision towards a more reliable outcome.}
\label{fig:method_overview}
\end{figure*}

\subsection{Overview of \name{}}
\label{approch:overview}

To mitigate error propagation during autoregressive generation, we introduce a novel \textit{Self-Reflective Generation at Test Time} (\name{}) framework. \name{} embeds a lightweight monitor-reflect-optimize loop into the autoregressive decoding process. This loop enables the model to identify and correct potential errors at intermediate steps, thereby mitigating their propagation throughout the generated sequence. As illustrated in Figure~\ref{fig:method_overview}, the process at each generation step \(t\) consists of two stages:

\textbf{Stage 1: Dynamic Uncertainty Monitoring.} At each step, the framework assesses the model predictive uncertainty for the next token. We quantify this uncertainty using token entropy. An intervention is triggered if this entropy exceeds a dynamic threshold that adapts to the local context of the generated sequence.

\textbf{Stage 2: Self-Reflective Optimization.} If the uncertainty exceeds the threshold, the standard decoding process is paused to initiate a self-reflective optimization. This optimization computes a transient correction vector \( \delta \) by minimizing a self-reflection loss function. The correction vector \( \delta \) is then applied to the current hidden state to refine the next-token probability distribution, guiding it toward a more confident and contextually coherent output. If the threshold is not met, the model proceeds with standard decoding.

The complete workflow is formalized in Algorithm~\ref{alg:srgen}. By selectively applying this real-time intervention, \name{} enhances output reliability with minimal and bounded computational overhead.

\subsection{Stage 1: Dynamic Uncertainty Monitoring}

A primary challenge in identifying critical reasoning junctures is that a fixed uncertainty threshold is suboptimal. Models with different architectures, training paradigms, or scales have distinct entropy profiles, even on the same task, as shown in Appendix~\ref{section:entropy_analysis}. A static threshold would therefore fail to reliably detect moments of high uncertainty across diverse contexts. To address this issue, we propose a dynamic thresholding strategy that adapts to the recent generation history of a model. Formally, at each decoding step \(t\), before sampling the next token \(y_t\), we first compute the predictive entropy of the next-token distribution given the prefix \(y_{<t} = (y_0, \dots, y_{t-1})\):
\begin{equation}
    H_t = H(p(\cdot | y_{<t})).
\end{equation}
We maintain a sliding window \(\mathcal{H}_t\) containing the \(N\) most recent entropy values, \(\{H_{t-N}, \dots, H_{t-1}\}\). From this history, we compute the running mean \(\mu(\mathcal{H}_t)\) and standard deviation \(\sigma(\mathcal{H}_t)\). The self-reflection process is triggered if the current entropy \(H_t\) represents a statistically significant deviation from the recent trend. Formally, reflection is activated if:
\begin{equation}
    H_t > \mu(\mathcal{H}_t) + k \cdot \sigma(\mathcal{H}_t),
\end{equation}
where \(k\) is a sensitivity hyperparameter. This adaptive approach enables our method to distinguish between naturally high-entropy passages and anomalous uncertainty spikes that warrant intervention. We present in the Section ~\ref{sec:critical_token} the tokens identified by this method.

\subsection{Stage 2: Self-Reflective Optimization}
\label{section:stage}

Once a high-uncertainty juncture is identified at step \(t\), our goal is to compute a transient correction to steer the generative process of a model. 

Just as humans pause to deliberate when facing indecision, eventually settling on the most prudent answer, our intuitive goal is for the model to reduce uncertainty at the current step and make a decisive choice, rather than sampling from an ambiguous distribution. However, blindly minimizing entropy carries a risk: it can cause the output distribution to collapse onto high-frequency tokens that are statistically dominant but contextually vacuous. Human reasoning typically involves a retrospective review of established thoughts, serving as a foundation for further deduction. Consequently, any reduction in uncertainty must be predicated on strict fidelity to the already-generated context. Therefore, a self-reflective generation process should inherently address two dual objectives: sharpening the predictive distribution to reduce uncertainty while preserving the semantic coherence established by the preceding context \(y_{<t}\). 

Inspired by the work of \citep{hu2025slot}, we introduce a transient correction vector \( \delta \in \mathbb{R}^d \), where \(d\) is the dimension of the model hidden state. This vector is initialized to zero and optimized only when the uncertainty monitor is triggered. The correction is applied to the final hidden state \(h_{t-1}\) before the vocabulary projection head \(\mathcal{W}\), yielding a modified logits vector:
\begin{equation}
    \label{equation:logits}
    \text{logits}_t' = \mathcal{W}(h_{t-1} + \delta).
\end{equation}
The optimization of \(\delta\) is guided by a hybrid loss function \(\mathcal{L}_{\text{SRGen}}\) defined over the prefix \(y_{<t}\):
\begin{equation}\label{eq:loss}
\begin{split}
    \mathcal{L}_{\text{SRGen}}(\delta; \lambda, y_{<t}) &= (1-\lambda)\mathcal{L}_{\text{CE}}(y_{<t}; \delta) \\
    &\quad + \lambda\mathcal{L}_{\text{AEM}}(y_{<t}; \delta).
\end{split}
\end{equation}
This loss comprises two components:
% \begin{itemize}[topsep=0pt,leftmargin=8mm,itemsep=0pt]
    % \item 

\noindent \textbf{\circled{1} Retrospective Context Loss (\(\mathcal{L}_{\text{CE}}\)).} This term ensures contextual fidelity by penalizing corrections \(\delta\) that disrupt the model predictions for the already-generated prefix. It is the negative log-likelihood of the prefix, where the same correction \(\delta\) is applied to all historical hidden states:
    \begin{equation}
        \mathcal{L}_{\text{CE}}(y_{<t};\delta) = -\sum_{i=0}^{t-2} \log p(y_{i+1} | y_{\le i}, \delta),
    \end{equation}
    where \( p(y_{i+1}|y_{\le i},\delta) = \text{softmax}(\mathcal{W}(h_{i}+\delta))_{y_{i+1}} \) and \(h_i\) is the hidden state corresponding to the prefix \(y_{\le i}\).

    % \item 
\noindent \textbf{\circled{2} Anticipatory Entropy Minimization (\(\mathcal{L}_{\text{AEM}}\)).} This term directly targets the high uncertainty at the current step \(t\). By minimizing the entropy of the next-token predictive distribution, it encourages the model to make a more confident prediction:
    \begin{equation}
        \mathcal{L}_{\text{AEM}}(y_{<t};\delta) = H(p(\cdot | y_{<t}, \delta)),
    \end{equation}
    where the perturbed distribution is \(p(\cdot | y_{<t}, \delta) = \text{softmax}(\mathcal{W}(h_{t-1} + \delta))\).
% \end{itemize}

After optimizing \(\delta\) for a few gradient steps to find \(\delta^*\), we use this correction to generate the token \(y_t\). The vector is then discarded, ensuring that each intervention is localized to its specific context.

\section{Method Analysis and Insights}
This section provides a deeper analysis of the \name{} framework, justifying its core design principles and theoretical underpinnings.

\subsection{Rationale for Dynamic and Selective Intervention}
The design of our uncertainty monitor is based on the principle of targeted intervention, which is crucial for two reasons: \textbf{(1) Efficiency:} Limiting the intervention to a few key tokens significantly reduces computational overhead. Applying reflection at every step would be computationally prohibitive, whereas our targeted approach adds only a modest, bounded overhead by focusing resources on the most critical junctures. \textbf{(2) Quality:} Excessive self-reflection can be counterproductive. Unnecessary intervention on low-uncertainty tokens may disrupt the fluency of the generated text. Our selective strategy, guided by the dynamic threshold, ensures that intervention is not only efficient but also beneficial to the final output quality.

\subsection{Theoretical Basis of the Hybrid Loss}

Our central theoretical result establishes that the hybrid loss in \name{} is not a heuristic but emerges directly from a principled optimization problem. We formalize this by showing that our loss function is equivalent to the Lagrangian of a constrained objective that seeks to minimize uncertainty while preserving contextual fidelity. This is stated formally in Theorem~\ref{thm:loss_equivalence} and a detailed proof is provided in Appendix~\ref{app:proof_loss_equivalence}.
\begin{tcolorbox}[colback=blue!3!white,colframe=blue!80!black,left=0.5mm,right=1mm,top=1mm,bottom=1mm]
\begin{theorem}[Hybrid Loss as Principled Constrained Optimization]
\label{thm:loss_equivalence}
\textit{\fontfamily{ppl}\selectfont
Given a trade-off parameter $\lambda \in (0,1)$, the minimizer $\delta^*$ of the hybrid loss objective
\begin{equation}
\label{eq:srgen_loss}
\mathcal{L}_{\text{SRGen}}(\delta; \lambda) = (1-\lambda)\,\mathcal{L}_{\text{CE}}(\delta) + \lambda\,\mathcal{L}_{\text{AEM}}(\delta),
\end{equation}
is also the solution to the constrained optimization problem
\begin{equation}
\label{eq:constrained_opt_formal}
\min_{\delta}\; \mathcal{L}_{\text{AEM}}(\delta) \quad \text{s.t.} \quad \mathcal{L}_{\text{CE}}(\delta) \le \epsilon.
\end{equation}
The choice of $\lambda$ implicitly defines the constraint boundary $\epsilon = \mathcal{L}_{\text{CE}}(\delta^*)$, establishing a formal equivalence between tuning the loss weight and setting a fidelity tolerance.
}
\end{theorem}
\end{tcolorbox}

\noindent\textbf{Remark.} This theorem provides a strong theoretical grounding for our method. The most powerful insight is that \textit{the \name{} objective is not an arbitrary blend of losses but a principled, tractable solution to a well-defined constrained optimization problem}. This reframes the intuitive goal of cautious generation, i.e., reducing future uncertainty ($\mathcal{L}_{\text{AEM}}$) without sacrificing fidelity to the current context ($\mathcal{L}_{\text{CE}} \le \epsilon$), in the rigorous language of optimization theory.

The parameter $\lambda$ is thus revealed to be more than a simple weighting factor; it implicitly controls the ``price'' of violating the contextual fidelity constraint. A small $\lambda$ (corresponding to a large Lagrange multiplier $\alpha$) enforces a strict fidelity requirement, heavily penalizing deviations. Conversely, a large $\lambda$ prioritizes uncertainty reduction, effectively relaxing the constraint. This perspective provides a formal basis for tuning $\lambda$. Furthermore, this result justifies our use of a simple weighted sum for the loss function. It demonstrates that this common practical approach is, in this case, equivalent to the more complex but formally correct Lagrangian relaxation, making the objective both theoretically sound and easily optimizable via standard gradient-based methods.

\subsection{Computational Overhead Analysis}

The computational overhead of \name{} comprises two components: a negligible monitoring stage and a more substantial, on-demand optimization stage. The monitoring stage, which computes predictive entropy at each token, incurs minimal cost because its inputs (logits and softmax distributions) are already computed during standard autoregressive generation.

Consequently, the primary overhead stems from the on-the-fly optimization of the correction vector $\delta$, which is triggered only at sparse, high-uncertainty junctures. This cost can be approximated as:
\begin{equation}
\text{Overhead} \approx N_{\text{act}} \times T \times C_{\text{bp}},
\label{eq:overhead}
\end{equation}
where $N_{\text{act}}$ is the number of reflection activations, $T$ is the number of inner optimization steps, and $C_{\text{bp}}$ is the cost of a single backpropagation pass. This design is inherently efficient. Unlike post-hoc refinement methods, whose costs scale linearly with the full sequence length, the overhead of our \name{} scales only with the number of critical interventions. Our experiments in Section~\ref{efficienct_result} showing that our method incurs only a small additional overhead. This makes \name{} a practical solution for enhancing model reasoning without prohibitive computational expense.

\section{Experiment}

\begin{table*}[t]
\centering
% 1. 设定字体大小：ACL 标准表格通常用 footnotesize 或 small
\footnotesize 

% 2. 调整行高：让表格不那么挤
\renewcommand{\arraystretch}{1.2} 

% 3. 调整列间距：
% 手动微调这个数值 (例如 4pt 到 6pt 之间)
% 让表格自然地接近页面宽度，但不要超出边界
\setlength{\tabcolsep}{5.5pt} 

\caption{Performance comparison on mathematical, general reasoning and code tasks. Best performance is bolded. Gray rows indicate SRGen.}
\label{tab:model_performance_main}

\begin{tabular}{
  l 
  l 
  c c c c % Math
  c c     % General
}
\toprule
& & \multicolumn{4}{c}{\textbf{Mathematical Reasoning}} & \multicolumn{2}{c}{\textbf{General \& Code}} \\
\cmidrule(lr){3-6} \cmidrule(lr){7-8} 
\textbf{Model} & \textbf{Method} & 
\textbf{AIME24} & \textbf{AIME25} & \textbf{HMMT25} & \textbf{AMC} & 
\textbf{GPQA} & \textbf{EvalPlus} \\
\midrule

% --- Qwen2.5-Math-7B ---
\multirow{3}{*}{\textbf{Qwen2.5-Math-7B}}
& CoT & 14.6 & 6.0 & 1.3 & 34.0 & 31.8 & 45.7 \\
& Self-Refine & 15.3\deltaT{+0.7} & 5.7\deltaT{-0.3} & 0.0\deltaT{-1.3} & 32.4\deltaT{-1.6} & 31.5\deltaT{-0.3} & 43.9\deltaT{-1.8} \\
\rowcolor{srgengray} 
\cellcolor{white} & SRGen & \best{22.0}\deltaT{+7.4} & \best{9.3}\deltaT{+3.3} & \best{3.3}\deltaT{+2.0} & \best{41.2}\deltaT{+7.2} & \best{33.8}\deltaT{+2.0} & \best{47.6}\deltaT{+1.9} \\
\midrule

% --- DeepSeek-R1-Distill-Qwen-7B ---
\multirow{3}{*}{\shortstack[l]{\textbf{DS-R1-Qwen-7B}}}
& CoT & 49.3 & 35.3 & 15.3 & 51.0 & 50.0 & 73.1 \\
& Self-Refine & 52.0\deltaT{+2.7} & 38.0\deltaT{+2.7} & 14.7\deltaT{-0.6} & \best{51.6}\deltaT{+0.6} & 51.2\deltaT{+1.2} & 72.6\deltaT{-0.5} \\
\rowcolor{srgengray}
\cellcolor{white} & SRGen & \best{61.3}\deltaT{+12.0} & \best{42.7}\deltaT{+7.4} & \best{19.3}\deltaT{+4.0} & 51.2\deltaT{+0.2} & \best{51.3}\deltaT{+1.3} & \best{73.7}\deltaT{+0.6} \\
\midrule

% --- DeepSeek-R1-Distill-Llama-8B ---
\multirow{3}{*}{\shortstack[l]{\textbf{DS-R1-Llama-8B}}}
& CoT & 48.0 & 30.7 & 14.0 & 50.0 & 46.4 & 71.3 \\
& Self-Refine & 48.7\deltaT{+0.7} & 32.0\deltaT{+1.3} & 16.0\deltaT{+2.0} & 49.4\deltaT{-0.6} & 47.9\deltaT{+1.5} & \best{73.1}\deltaT{+1.8} \\
\rowcolor{srgengray}
\cellcolor{white} & SRGen & \best{53.3}\deltaT{+5.3} & \best{36.0}\deltaT{+5.3} & \best{18.0}\deltaT{+4.0} & \best{50.6}\deltaT{+0.6} & \best{48.5}\deltaT{+2.1} & 72.0\deltaT{+0.7} \\
\midrule

% --- Qwen3-32B ---
\multirow{3}{*}{\textbf{Qwen3-32B}}
& CoT & 76.7 & 70.7 & 23.3 & 54.0 & 63.5 & 87.1 \\
& Self-Refine & 76.7\deltaT{+0.0} & 69.3\deltaT{-1.4} & 25.3\deltaT{+2.0} & 54.4\deltaT{+0.4} & 64.9\deltaT{+1.4} & 87.2\deltaT{+0.1} \\
\rowcolor{srgengray}
\cellcolor{white} & SRGen (Ours) & \best{82.7}\deltaT{+6.0} & \best{76.0}\deltaT{+5.3} & \best{28.0}\deltaT{+4.7} & \best{56.8}\deltaT{+2.8} & \best{65.7}\deltaT{+2.2} & \best{87.8}\deltaT{+0.7} \\
\bottomrule
\end{tabular}
\end{table*}

\subsection{Experimental Setup}

\textbf{Models.} 
We assess generality on open-weight models spanning scales, architectures, and post-training regimes: \texttt{Qwen2.5-Math-7B}~\citep{team2024qwen2}, \texttt{DeepSeek-R1-Distill-Qwen-7B}, \texttt{DeepSeek-R1-Distill-Llama-8B}~\citep{guo2025deepseek}, and \texttt{Qwen3-32B}~\citep{yang2025qwen3}. This set covers two architecture families (\texttt{Qwen}, \texttt{Llama}), sizes from 7B to 32B, and heterogeneous post-training pipelines (distillation, SFT, RL) that yield distinct entropy profiles. 
% Open weights ensure reproducibility and permit the \textit{hidden-state} access required for our test-time intervention. 
This diversity probes whether \name{} remains effective across decoding behaviors, tokenizers, and training regimes, rather than overfitting to any single family or size.

\textbf{Benchmarks.} 
We evaluate on \texttt{AIME2024}, \texttt{AIME2025}~\citep{AoPSWikiAIME}, \texttt{HMMT2025}~\citep{balunovic_srimatharena_2025}, and \texttt{AMC}~\citep{numina_math_datasets} for mathematical reasoning problems, \texttt{GPQA}~\citep{rein2024gpqa} for general reasoning and \texttt{EvalPlus}~\citep{liu2023your} for code generation. These tasks typically require accurate answers, so small early slips tend to propagate and can overturn the final result.

\textbf{Experimental Settings.} 
For inference, we cap the maximum generation length at 4{,}096 tokens for \texttt{Qwen2.5-Math-7B} and 32{,}768 for all other models. Decoding uses temperature $0.6$ with nucleus sampling at top-$p=0.95$, following common recommendations for reasoning models. We also report \texttt{Qwen2.5-Math-7B} performance at temperature $0$ in subsequent analyses. All experiments were conducted on NVIDIA A800-80G GPUs. Unless otherwise noted, our method uses the following hyperparameters: inner update steps $T=3$, learning rate $\eta=0.01$, entropy-detection window $N=25$, and standard-deviation multiplier $k=4$. We conduct hyperparameter ablation studies in Appendix~\ref{section:ablation}.

\textbf{Metrics.}
Our accuracy is based on five runs of pass@1.

\subsection{Main Results}

In Table~\ref{tab:model_performance_main}, we compare the improvements brought by Self-Refine and \name{} on CoT-style reasoning. The results show that \name{} consistently achieves stronger performance across mathematical, general reasoning, and coding tasks, and this trend holds across models trained under different paradigms. Notably, on \texttt{AIME2024}, \name{} improves the accuracy of \texttt{DeepSeek-Distill-Qwen-7B} by 12 percentage points and \texttt{Qwen3-32B} by 6 percentage points, substantially larger than the gains obtained by Self-Refine on this type of task.

These results suggest that, even without any external feedback, \name{} can leverage a self-reflective generation process to reduce reasoning errors during decoding. By mitigating cascading mistakes and preventing early errors from propagating into irreversible failure modes, \name{} yields more reliable intermediate reasoning and ultimately improves the model’s overall reasoning capability.

\begin{table*}[h]
\centering
\footnotesize
\renewcommand{\arraystretch}{1.2}
\setlength{\tabcolsep}{4.2pt}

\caption{Efficiency comparison on 100 \texttt{MATH500} problems using \texttt{Qwen2.5-Math-7B}. We report wall-clock time, generated tokens, peak memory usage, and average time per example. Lower is better.}
\label{tab:efficiency_comparison}

\begin{tabular}{l c c c c}
\toprule
\textbf{Method} & 
\textbf{\shortstack[c]{Wall-clock Time (s)}} & 
\textbf{\shortstack[c]{Tokens Generated (w)}} & 
\textbf{\shortstack[c]{Memory Peak (GB)}} & 
\textbf{\shortstack[c]{Avg Time per example (s)}} \\
\midrule
Base        & 1025 & 7.2  & 15.7 & 10.2 \\
Self-Refine & 2316 & 15.6 & 18.7 & 23.2 \\
SLOT        & 1148 & 7.5  & 18.5 & 11.5 \\
MI-Peak     & 1744 & 12.1 & 16.6 & 17.4 \\
\rowcolor{srgengray}
SRGen       & 1198 & 8.0  & 17.8 & 12.0 \\
\bottomrule
\end{tabular}
\end{table*}

\begin{figure}[h]
  \centering
  \includegraphics[width=\columnwidth]{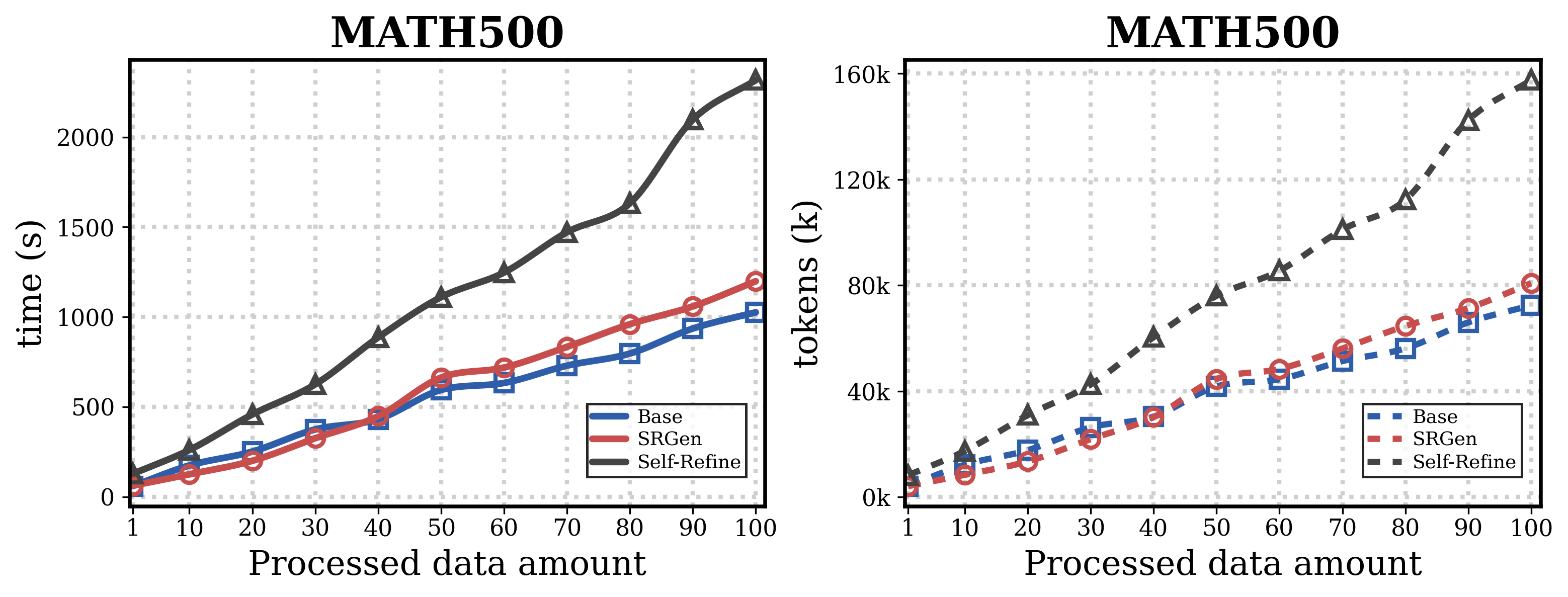}
  \caption{Time increase and token cost.}
  \label{fig:wrap_left}
\end{figure}

\begin{figure*}[t]
    \centering
    \includegraphics[width=\textwidth]{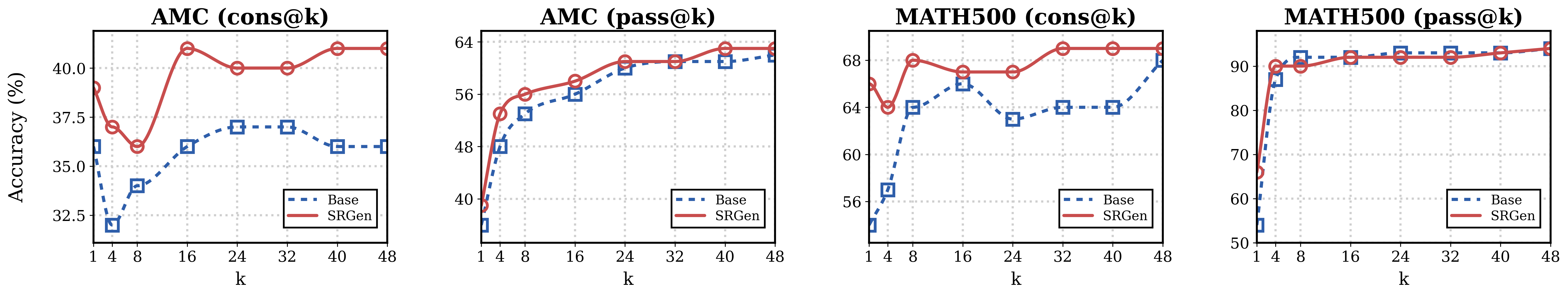}
    \caption{\texttt{Cons@k} and \texttt{Pass@k} accuracy of \texttt{Qwen2.5\mbox{-}Math\mbox{-}7B} on \texttt{AMC} and \texttt{MATH500}}
\label{fig:cons_pass_k}
\end{figure*}

In addition, we directly compare SRGen with several methods that also intervene during the generation process. To align with the experimental setup of MI-Peak, we use greedy decoding for the math tasks. The results show that SRGen yields more stable improvements in most cases. Table ~\ref{tab:greedy_analysis_updated} reports detailed results.

\begin{table}[h]
\centering
\footnotesize 
\renewcommand{\arraystretch}{1.2} 
\setlength{\tabcolsep}{4.5pt} 

\caption{Deterministic performance comparison using Greedy Decoding.}
\label{tab:greedy_analysis_updated}

\begin{tabular}{
  l 
  l 
  c c c 
}
\toprule
\textbf{Model} & \textbf{Method} & 
\textbf{MATH} & \textbf{AIME} & \textbf{AMC} \\
\midrule

% --- Qwen2.5-1.5B ---
\multirow{4}{*}{\shortstack[l]{\textbf{Qwen2.5-3B-}\\\textbf{Instruct}}}
& CoT & 62.6 & 10.0 & 39.0 \\ 
& MI-Peak & 63.8 & 6.7 & 41.0 \\
& SLOT & 66.2 & 10.0 & 40.0 \\
\rowcolor{srgengray}
\cellcolor{white} & SRGen & \best{66.6} & \best{10.0} & \best{43.0} \\
\midrule

% --- Qwen2.5-7B ---
\multirow{4}{*}{\shortstack[l]{\textbf{Qwen2.5-Math-}\\\textbf{7B}}}
& CoT & 63.8 & 13.3 & 35.0 \\
& MI-Peak & 64.0 & 10.0 & 35.0 \\
& SLOT & 64.2 & 20.0 & 35.0 \\
\rowcolor{srgengray}
\cellcolor{white} & SRGen & \best{69.4} & \best{20.0} & \best{37.0} \\
\midrule

% --- Llama-8B ---
\multirow{4}{*}{\textbf{DS-R1-}\textbf{Qwen-7B}}
& CoT & 85.6 & 40.0 & 48.0 \\
& MI-Peak & 87.0 & 50.0 & 46.0 \\
& SLOT & 85.0 & 46.7 & \best{49.0} \\
\rowcolor{srgengray}
\cellcolor{white} & SRGen & \best{87.2} & \best{53.3} & 48.0 \\
\midrule

% --- DeepSeek-R1-Distill-Qwen-7B ---
\multirow{4}{*}{\shortstack[l]{\textbf{DS-R1-}\textbf{Llama-8B}}}
& CoT & 80.4 & 43.3 & 42.0 \\
& MI-Peak & 81.8 & 40.0 & 43.0 \\
& SLOT & 84.4 & 43.3 & 45.0 \\
\rowcolor{srgengray}
\cellcolor{white} & SRGen & \best{84.8} & \best{46.7} & \best{46.0} \\
\bottomrule
\end{tabular}
\end{table}

\subsection{Efficiency Analysis}
\label{efficienct_result}

We evaluate inference efficiency after integrating \name{} by measuring both wall-clock runtime and token usage. Experiments are conducted with \texttt{Qwen2.5-Math-7B} on \texttt{MATH500}. We report the mean per-task runtime and the total number of generated tokens, and we use greedy decoding for all settings to remove randomness. Figure~\ref{fig:wrap_left} shows results as we vary the processed data amount. Across the full range, \name{} closely tracks the baseline in both runtime and token consumption. The time overhead remains limited and increases smoothly with the processed data amount, while the token count stays comparable to the base model. In contrast, Self-Refine incurs substantially higher cost, with noticeably steeper growth in both runtime and tokens as the processed data amount increases. These results indicate that \name{} improves performance while maintaining efficient inference, and it avoids the large efficiency degradation observed with Self-Refine.

By evaluating the efficiency of different methods on 100 \texttt{MATH500} problems using \texttt{Qwen2.5-Math-7B}, we also provide a more detailed comparison, including wall-clock time, tokens generated, and peak memory usage, as shown in Table~\ref{tab:efficiency_comparison}. The results indicate that, even compared with test-time methods, \name{} still maintains relatively high efficiency while delivering stronger performance.

\subsection{Cons@k and Pass@k}

\begin{figure*}[t]
    \centering
    \includegraphics[width=\textwidth]{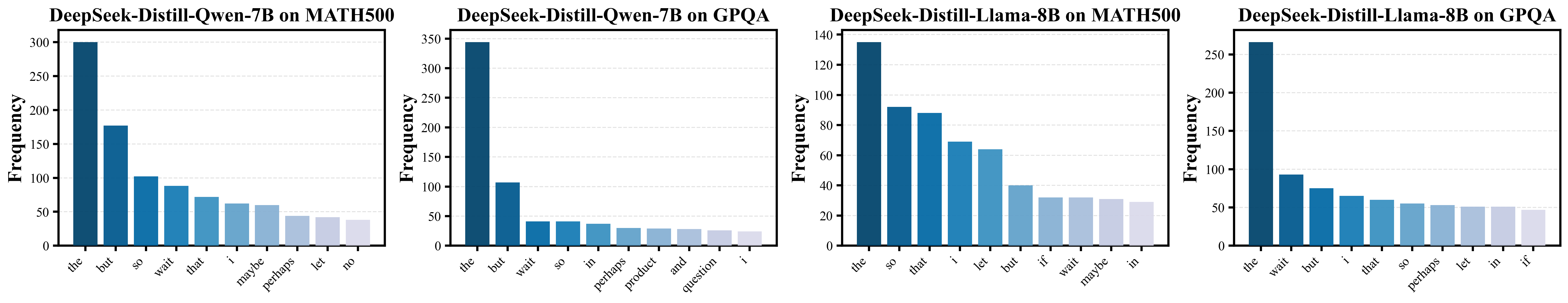}
    \caption{Critical tokens identified by the dynamic entropy monitor.}
\label{fig:token_freq}
\end{figure*}

Using \texttt{Qwen2.5\mbox{-}Math\mbox{-}7B} on \texttt{AMC} and \texttt{MATH500}, we further examine how \texttt{Cons@k} and \texttt{Pass@k} change with increasing $k$ after applying \name{}. Detailed results are shown in Figure~\ref{fig:cons_pass_k}. When \name{}-enhanced samples are used for self-consistency voting, \texttt{Cons@k} remains consistently higher than the base model as $k$ grows, whereas \texttt{Pass@k} gradually converges to the base model. These trends indicate that self-reflective generation in \name{} reduces the probability of reasoning errors and improves single-pass accuracy, thereby producing higher-quality candidates for self-consistency and boosting \texttt{Cons@k}. Importantly, the reflection mechanism in \name{} does not diminish the exploratory capacity of the model: it primarily reduces mistakes by reweighting uncertain tokens, correcting erroneous traces while leaving correct ones intact. Taken together, these results suggest that \name{} is a promising plug-in for achieving higher single-pass accuracy or more sample-efficient self-consistency.

\subsection{Orthogonality: Integration with Related Methods}
\label{section:orth}

\begin{figure}[h]
\begin{center}
  \includegraphics[width=\columnwidth]{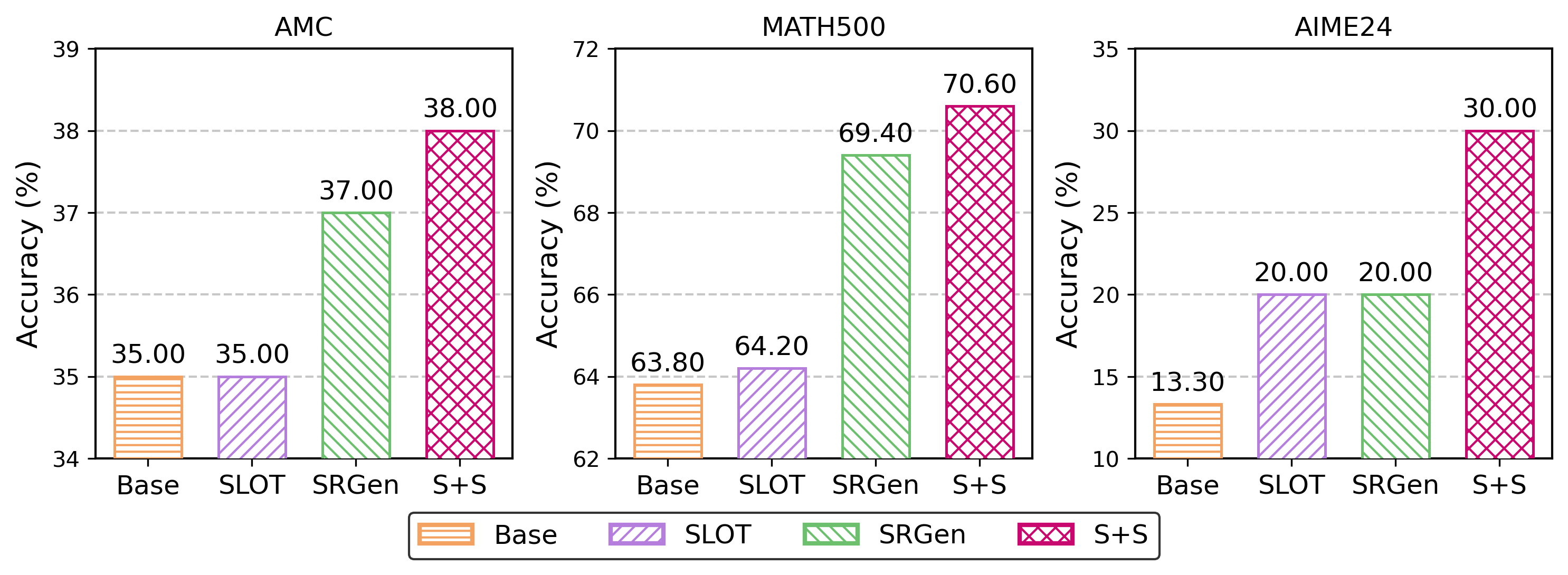}
\end{center}
\caption{Performance of SLOT, \name{}, and their combination for \texttt{Qwen2.5\mbox{-}Math\mbox{-}7B} across the \texttt{AMC}, \texttt{MATH500}, and \texttt{AIME2024} benchmarks.}
\label{fig:compare_slot_\name{}}
\end{figure}

Building on our initial results showing that \name{} can pair with diverse training paradigms and further improve models post-training, we next provide stronger evidence of its orthogonality and potential synergy by combining it with a method from the same family. Specifically, we adopt SLOT, a representative test-time optimization approach. During inference, SLOT optimizes a sample-specific vector $\delta_{\text{SLOT}}$ over the prompt-processing stage and injects it into the hidden states of the model to steer generation. All experiments are conducted on the \texttt{Qwen2.5-Math-7B} model, and we use greedy decoding throughout to ensure reproducibility.

As shown in Figure~\ref{fig:compare_slot_\name{}}, combining SLOT with \name{} further strengthens the reasoning ability of the model, with particularly pronounced gains on mathematical tasks. On \texttt{MATH500}, the joint approach lifts \texttt{Qwen2.5\mbox{-}Math\mbox{-}7B} from 63.8\% to 70.6\%, outperforming either SLOT or \name{} used in isolation. On \texttt{AMC}, it improves performance from 35.0\% to 38.0\%, exceeding both SLOT (35.0\%) and \name{} (37.0\%). The synergy is even more evident on \texttt{AIME2024}, where the combined method boosts accuracy from 13.3\% to 30.0\%, substantially outperforming either SLOT or \name{} alone (both at 20.0\%). These results provide additional evidence that \name

\subsection{Critical Token Identified}
\label{sec:critical_token}
In this section, we analyze which tokens are flagged by our dynamic entropy monitor. We run \texttt{DeepSeek-R1-Distill-Qwen-7B} and \texttt{DeepSeek-R1-Distill-Llama-8B} on \texttt{MATH500} and \texttt{GPQA}, and collect tokens whose next-token uncertainty exceeds an adaptive threshold. Figure~\ref{fig:token_freq} reports the aggregated frequencies.

Across both models and benchmarks, the most frequent tokens are mainly function words and discourse connectives, including \texttt{the}, \texttt{so}, \texttt{but}, \texttt{that}, \texttt{perhaps}, \texttt{maybe}, \texttt{let}, \texttt{if}, and \texttt{for}. These tokens often appear near clause boundaries and decision points in the chain of thought, where the model chooses between plausible continuations such as introducing a new premise, changing direction, or committing to the next step. As a result, uncertainty spikes are more likely to occur on tokens that guide the reasoning trajectory, rather than on content tokens that only add details.

This pattern supports our design choice. A simple dynamic threshold tends to surface high-impact connective tokens, which provides meaningful intervention points without spending computation on low-value positions. In practice, concentrating inner updates on this compact set keeps optimization stable and efficient. Early in the updates, the cross-entropy term and the entropy-reduction term often push in similar directions, improving both confidence and consistency with the current context. As updates continue, the two objectives may start to conflict. The weight $\lambda$ then controls the trade-off between fitting the current context and reducing uncertainty at each triggered step.

Overall, Figure~\ref{fig:token_freq} suggests that entropy-guided, token-aware intervention targets the specific junctures that most strongly influence downstream reasoning. This helps explain the consistent gains we observe across models and benchmarks, while keeping the overhead modest and bounded.

\section{Conclusion}

We introduce \textit{Self-Reflective Generation at Test Time} (\name{}), a lightweight, plug-and-play framework that performs token-level self-reflection only at critical tokens detected by \textit{dynamic uncertainty monitoring}. When triggered, a brief \textit{self-reflective optimization} learns an on-the-fly correction vector \(\delta\) and injects it into the hidden state under a hybrid loss \(L_{\text{SRGen}}\) to reduce predictive uncertainty while preserving contextual fidelity. Across challenging benchmarks spanning mathematical reasoning, general reasoning, and code generation, \name{} improves accuracy with small additional inference time and yields more effective self-consistency voting without harming exploration. \name{} is practical: it strengthens reasoning and generation across model architectures and training paradigms and composes with other test-time methods, making it a promising inference-time plug-in for more reliable LLM generation. 

Moreover, the effectiveness of \name{} further validates the importance of critical tokens in the model’s reasoning process, suggesting that appropriately intervening on these tokens can effectively improve reasoning performance

\section*{Limitations}

In this section, we discuss several potential limitations of \name{}. First, SRGen intervenes during generation based on an uncertainty signal and performs online optimization of the corrective vector at each trigger point. This typically requires access to the model's hidden states and gradient information. As a result, \name{} is most suitable for controllable open-weight or white-box inference settings, and it may be difficult to deploy in scenarios that only provide a text-only interface or a restricted inference stack. Second, \name{} relies on a dynamic threshold to identify uncertain positions. As shown in Section~\ref{sec:critical_token}, this strategy can effectively capture key junctions in the reasoning process, and intervening at these positions leads to consistent gains. However, the current method for identifying such critical points still has substantial room for improvement. Future work could explore stronger position-selection strategies so that \name{} can target more precise and informative locations. Third, our method introduces several additional hyperparameters. In Appendix ~\ref{section:ablation}, we show that \name{} is relatively stable across different hyperparameter choices and we provide a general default configuration. Nevertheless, the hyperparameters that yield the best performance may vary across models and tasks. For settings that require peak performance, future work may study efficient tuning on a small validation set, or develop adaptive hyperparameter schemes to improve the applicability of \name{}.

\section*{Ethical Considerations}
Our method aims to incorporate self-reflective decision making into the model's generation process and can be easily integrated as a test-time technique. It does not create new datasets or train new models and introduce additional risks related to discrimination, bias, exploitation, human rights violations, or other ethical concerns.

\bibliography{custom}

\clearpage

\appendix

\section{Related Work}

\textbf{Self-Reflection in LLMs.}
Self-reflection seeks to move LLMs from impulsive first-pass outputs to more deliberative and accurate responses. Existing approaches largely fall into two categories. (1) Post hoc iterative refinement. These methods use multi-stage inference pipelines: the model first drafts an answer, then critiques it, and finally revises it based on its own feedback \citep{ shinn2023reflexion, yu2024teaching, yuksekgonul2024textgrad, li2025test}. Frameworks such as Self-Refine \citep{madaan2023self} formalize the generate, critique, and refine loop. While effective, they incur substantial latency and computational overhead because they require multiple full forward passes. (2) Training for intrinsic self-correction. This line embeds self-correction directly in the parameters, typically via fine-tuning on corrective data or reinforcement learning (RL) \citep{bai2022constitutional, hu2024teaching,  kumar2024training, moskvoretskii2025self}. Such models can produce refined outputs in a single pass but demand extensive and costly training.
\name{} offers a distinct alternative that operates at test time. It avoids the high overhead of iterative methods by requiring neither auxiliary LLM calls for feedback nor the generation of multiple complete outputs. Operating at the token level. \name{} is a lightweight method and can be synergistically combined with both training-based and post-hoc self-reflection approaches.

\textbf{Identifying and Leveraging Critical Tokens.}
Recent work rests on the observation that tokens are not equally informative \citep{lincritical}. Studies identify “critical” or “pivotal” tokens that mark decision points along a reasoning path and leverage them in several ways. (1) Guiding training: policy gradients are applied selectively at high-entropy positions to focus learning \citep{wang2025beyond, vassoyan2025ignore}. (2) Triggering exploration: critical tokens act as branching points for sampling diverse reasoning paths \citep{zheng2025first, zhu2025uncertainty} or for localized iterative refinement that probes the solution space more deeply \citep{qian2025demystifying}. (3) Pruning search: low-confidence tokens trigger the removal of less promising paths within self-consistency frameworks \citep{fu2025deep, taubenfeld2025confidence, zhou2025bridging}.
We introduce a new paradigm for the use of critical tokens. \name{} employs them as real-time triggers for a corrective intervention directly on the model's hidden state. This allows for an on-the-fly steering of a single generation path, a fundamentally different and more direct mechanism than prior art.

\textbf{Test-Time Scaling.} To improve performance without costly retraining, a range of methods increase computation at test time. These approaches fall into two broad strategies. The first generates multiple reasoning paths and selects an outcome via voting or scoring. This includes producing multiple complete solutions, as in self-consistency \citep{wang2022self, singhi2025solve}, or exploring a tree or graph of intermediate steps, as in Tree-of-Thoughts and its variants \citep{yao2023tree, bi2024forest, teng2025atom}. The second intervenes within a single decoding process. Prompt-based techniques such as chain of thought \citep{wei2022chain} elicit more deliberative reasoning. More directly, methods adjust the model’s internal computations during a single pass, e.g., DoLa \citep{chuang2023dola} contrasts layer logits to steer decoding and SLOT \citep{hu2025slot} injects a sample-specific vector into the hidden states to steer generation globally and indirectly encourages longer reasoning by suppressing the EOS token.
\name{} advances the second strategy with a fine-grained, dynamic intervention. Whereas SLOT applies a static, sample-level vector throughout decoding, \name{} computes a token-level corrective vector $\delta$ on-the-fly at detected critical junctures. This targeted adjustment adapts to the immediate context and steers the reasoning process without branching or multiple full passes.

\section{More Results}

\section{Theorems and proofs}
\subsection{Proof of Thm.~\ref{thm:loss_equivalence}}
\label{app:proof_loss_equivalence}

\textbf{Statement.}
Fix a trade-off parameter $\lambda\in(0,1)$. Let
\[
F_\lambda(\delta)
\;\triangleq\;
(1-\lambda)\,L_{\text{CE}}(\delta)\;+\;\lambda\,L_{\text{AEM}}(\delta),
\]
where $L_{\text{CE}}$ and $L_{\text{AEM}}$ are defined in the main text
(\S~\ref{section:stage}) on the current prefix $y_{<t}$ with the same correction
vector $\delta$ injected as in Eq.~\ref{equation:logits}.
Let $\delta^\star \in \arg\min_\delta F_\lambda(\delta)$ be any minimizer.
Then $\delta^\star$ also solves the constrained problem
\begin{equation}
\min_{\delta}\; L_{\text{AEM}}(\delta)
\quad\text{s.t.}\quad
L_{\text{CE}}(\delta)\;\le\;\varepsilon,
\label{eq:cons}
\end{equation}
with the \emph{implicitly induced} tolerance
$\varepsilon\;\triangleq\;L_{\text{CE}}(\delta^\star)$.

\textbf{Proof.}
Let $\varepsilon=L_{\text{CE}}(\delta^\star)$. By construction,
$\delta^\star$ is feasible for \eqref{eq:cons} with the constraint held
at equality. Suppose, for contradiction, that $\delta^\star$ is not an
optimizer of \eqref{eq:cons}; then there exists a feasible
$\widehat{\delta}$ with $L_{\text{CE}}(\widehat{\delta})\le\varepsilon$
and $L_{\text{AEM}}(\widehat{\delta})<L_{\text{AEM}}(\delta^\star)$.
Consider the hybrid objective values:
\begin{equation}
\begin{aligned}
F_\lambda(\widehat{\delta})
&= (1-\lambda)L_{\text{CE}}(\widehat{\delta})
   + \lambda L_{\text{AEM}}(\widehat{\delta}) \\
&\le (1-\lambda)\varepsilon
   + \lambda L_{\text{AEM}}(\widehat{\delta}) \\
&<  (1-\lambda)\varepsilon
   + \lambda L_{\text{AEM}}(\delta^\star) \\
&= F_\lambda(\delta^\star).
\end{aligned}
\label{eq:chain}
\end{equation}

which contradicts the optimality of $\delta^\star$ for
$F_\lambda$. Hence no feasible point can achieve a strictly smaller
$L_{\text{AEM}}$ under the tolerance $L_{\text{CE}}\le\varepsilon$,
and $\delta^\star$ solves \eqref{eq:cons}. 

\textbf{Lagrangian view and the $\lambda$--$\alpha$ mapping.}
The constrained problem \eqref{eq:cons} has Lagrangian
\[
\mathcal{L}(\delta,\alpha)
\;=\;
L_{\text{AEM}}(\delta)+\alpha\big(L_{\text{CE}}(\delta)-\varepsilon\big),
\qquad \alpha\ge 0.
\]
For any fixed $\alpha\ge 0$, minimizing $\mathcal{L}$ over $\delta$
is, up to an additive constant $-\alpha\varepsilon$, equivalent to
minimizing the \emph{weighted sum}
$L_{\text{AEM}}(\delta)+\alpha L_{\text{CE}}(\delta)$.
Identifying weights gives the bijection
\[
\lambda \;=\;\frac{1}{1+\alpha},
1-\lambda\;=\;\frac{\alpha}{1+\alpha},
\alpha\;=\;\frac{1-\lambda}{\lambda}.
\]
Therefore, the hybrid loss $F_\lambda$ is exactly a rescaled
Lagrangian with dual variable $\alpha=\frac{1-\lambda}{\lambda}$.
Under standard regularity ensuring KKT optimality (e.g., existence of
a primal optimum and either convexity with Slater’s condition or other
sufficient conditions for strong duality), any primal-dual optimal pair
$(\delta^\star,\alpha^\star)$ of \eqref{eq:cons} also minimizes a
weighted sum, and the mapping above recovers $\lambda$ from $\alpha^\star$.
This provides the converse direction under mild assumptions:
\emph{for a given active tolerance $\varepsilon$, an appropriate choice
of $\lambda$ (equivalently, $\alpha$) recovers the same optimizer
$\delta^\star$.}

\textbf{Pareto-optimality interpretation.}
Consider the bi-objective vector
$G(\delta) \triangleq \big(L_{\text{CE}}(\delta),\,L_{\text{AEM}}(\delta)\big)$.
By the proof above, any minimizer $\delta^\star$ of $F_\lambda$ with
$\lambda\in(0,1)$ is \emph{Pareto-optimal}: if there existed
$\widehat{\delta}$ with $G(\widehat{\delta})\preceq G(\delta^\star)$ and
one component strictly smaller, it would violate the optimality of
$\delta^\star$ for $F_\lambda$. Hence the hybrid loss selects points on
the Pareto front of the two desiderata “contextual fidelity” and
“uncertainty reduction.” In particular, the induced tolerance
$\varepsilon=L_{\text{CE}}(\delta^\star)$ characterizes the specific
frontier point attained.

\textbf{On the $\lambda\!\leftrightarrow\!\varepsilon$ trade-off.}
Intuitively, larger $\lambda$ increases the relative price on
$L_{\text{AEM}}$ and relaxes the pressure on $L_{\text{CE}}$, thus
tending to yield solutions with lower $L_{\text{AEM}}$ and (weakly)
higher $L_{\text{CE}}$ (i.e., a looser fidelity tolerance).
Formally, consider any $0<\lambda_1<\lambda_2<1$ with corresponding
minimizers $\delta_1,\delta_2$. Optimality implies
\[
F_{\lambda_1}(\delta_1)\le F_{\lambda_1}(\delta_2),
F_{\lambda_2}(\delta_2)\le F_{\lambda_2}(\delta_1).
\]
Writing $F_\lambda=(1-\lambda)L_{\text{CE}}+\lambda L_{\text{AEM}}$ and rearranging yields the \emph{weighted} trade-off bounds:

\begin{equation}
\label{eq:weighted_tradeoff}
\begin{aligned}
&\frac{1-\lambda_1}{\lambda_1}\big[L_{\text{CE}}(\delta_1)-L_{\text{CE}}(\delta_2)\big] \\
&\le\; L_{\text{AEM}}(\delta_2)-L_{\text{AEM}}(\delta_1) \\
&\le\; \frac{1-\lambda_2}{\lambda_2}\big[L_{\text{CE}}(\delta_1)-L_{\text{CE}}(\delta_2)\big].
\end{aligned}
\end{equation}

Thus the \emph{relative weights} $(1-\lambda)/\lambda$ govern the paired improvements: as $\lambda$ increases (placing more emphasis on $L_{\text{AEM}}$), the achievable decrease in $L_{\text{AEM}}$ per unit increase in $L_{\text{CE}}$ becomes tighter. In strictly convex or uniqueness regimes this typically induces a monotone path $\varepsilon(\lambda)=L_{\text{CE}}(\delta_\lambda)$ that is nondecreasing in $\lambda$.\footnote{We avoid global convexity claims for $L_{\text{AEM}}$; the sufficiency result above does not require convexity. In practice, a unique local minimizer selected by a deterministic inner optimizer makes the $\lambda$-path stable.}

\textbf{Boundary cases and feasibility.}
When $\lambda\to 1$ ($\alpha\to 0$), $F_\lambda$ approaches
$L_{\text{AEM}}$, i.e., the unconstrained entropy-minimization objective.
The induced tolerance becomes $\varepsilon=L_{\text{CE}}(\delta^\star)$
for an $L_{\text{AEM}}$-minimizer $\delta^\star$, so the constraint is
\emph{tight (active at equality)} in this mapping.
When $\lambda\to 0$ ($\alpha\to\infty$), $F_\lambda$ emphasizes
$L_{\text{CE}}$ and the solution tends to minimize contextual distortion
subject to making any progress on $L_{\text{AEM}}$; operationally this
corresponds to a nearly “hard” fidelity constraint.

\textbf{Existence of minimizers.}
Both $L_{\text{CE}}$ and $L_{\text{AEM}}$ in our setting are
nonnegative and continuous in $\delta$ (they are compositions of smooth
maps: affine shift in logits, softmax, entropy, and prefix NLL). Thus
$F_\lambda$ is lower-bounded by $0$ and continuous.
If $\arg\min F_\lambda$ fails to exist on $\mathbb{R}^d$ due to lack of
coercivity, it suffices (and is standard at test time) to either:
(i) restrict $\delta$ to a compact trust region
$\{\|\delta\|\le R\}$, or (ii) add a tiny quadratic regularizer
$\frac{\gamma}{2}\|\delta\|^2$ (this does not affect the equivalence to
\eqref{eq:cons} because the same regularizer can be added to both the
constrained and weighted formulations). Under either modification, a
minimizer exists and the above arguments apply verbatim.

\textbf{Joint-Descent Lemma.}
If the gradients form an acute angle, i.e.,
$\langle \nabla L_{\text{CE}}(\delta),\, \nabla L_{\text{AEM}}(\delta)\rangle>0$,
then for any $\lambda\in(0,1)$ and sufficiently small $\eta>0$, the update
\[
\delta^+\;=\;\delta-\eta\big[(1-\lambda)\nabla L_{\text{CE}}(\delta)+\lambda\nabla L_{\text{AEM}}(\delta)\big]
\]
strictly decreases \emph{both} objectives to first order.
\textit{Proof.} Directional derivatives give
\begin{equation}
\label{eq:deriv_neg}
\begin{aligned}
&\frac{d}{d\eta}L_{\text{CE}}(\delta^+)\big|_{\eta=0} \\
&= -\big[(1-\lambda)\|\nabla L_{\text{CE}}\|^2
       + \lambda\langle \nabla L_{\text{CE}}, \nabla L_{\text{AEM}}\rangle\big] \\
&< 0.
\end{aligned}
\end{equation}

\begin{equation}
\label{eq:aem_deriv_neg}
\begin{aligned}
&\frac{d}{d\eta}L_{\mathrm{AEM}}(\delta^+)\big|_{\eta=0} \\
&= -\big[\lambda\|\nabla L_{\mathrm{AEM}}\|^2
      +(1-\lambda)\langle \nabla L_{\mathrm{AEM}}, \nabla L_{\mathrm{CE}}\rangle\big] \\
&< 0.
\end{aligned}
\end{equation}
where both inequalities use the acute-angle assumption. \hfill$\square$

\textbf{Takeaway for SRGen.}
The proof establishes that the SRGen loss is not an ad-hoc blend: it is
precisely a Lagrangian relaxation of the constrained goal “reduce
uncertainty while keeping contextual fidelity within tolerance.”
Therefore $\lambda$ is an interpretable knob: it \emph{implicitly}
sets the fidelity tolerance $\varepsilon=L_{\text{CE}}(\delta^\star)$ and
moves SRGen along the fidelity--confidence Pareto frontier.
In practice (cf.\ Fig.~5 in the main text), small but nonzero $\lambda$
often works well, reflecting a regime where fidelity is enforced
strongly while still reaping entropy reductions at high-uncertainty
points.

\section{Algorithm of \name{}}

\begin{algorithm}[h]
\caption{\name{}: Self-Reflective Generation}
\label{alg:srgen}
\begin{algorithmic}[1]
\State \textbf{Input:} pre-trained model $M$ with head $W$; prompt $x_0$
\Statex \hspace{\algorithmicindent}\textit{Hyperparameters:} $k$ (sensitivity), $N$ (window size), $\lambda$ (loss weight), $T$ (steps), $\eta$ (lr), $\tau$ (temperature)
\State \textbf{Output:} generated sequence $y$

\State $y \gets ()$, $t \gets 1$, $\mathcal{E} \gets$ empty ring buffer of size $N$
\While{EOS not generated \textbf{and} $|y|<$ MAX\_LENGTH}
    \State $h_{t-1} \gets M(x_{0:t})$ \Comment{last hidden state for current context}
    \State $z \gets W\,h_{t-1}$; \quad $E_t \gets \mathrm{Entropy}(\mathrm{softmax}(z/\tau))$
    \If{$|\mathcal{E}|=N$ \textbf{and} $E_t > \mu(\mathcal{E}) + k\,\sigma(\mathcal{E})$} \Comment{dynamic trigger}
        \State $\delta \gets \mathbf{0}$
        \For{$i=1$ \textbf{to} $T$} \Comment{inner optimization of $\delta$}
            \State $\mathcal{L}_{\mathrm{CE}} \gets -\sum_{j=0}^{t-2}\log p(x_{j+1}\mid x_{0:j},\delta)$
            \State $\mathcal{L}_{\mathrm{AEM}} \gets -\sum_{v\in\mathcal{V}} p(v\mid x_{0:t},\delta)\log p(v\mid x_{0:t},\delta)$
            \State $\mathcal{L} \gets (1-\lambda)\,\mathcal{L}_{\mathrm{CE}} + \lambda\,\mathcal{L}_{\mathrm{AEM}}$
            \State $\delta \gets \delta - \eta\,\nabla_{\delta}\mathcal{L}$
        \EndFor
        \State $z \gets W\,(h_{t-1} + \delta)$ \Comment{modify only the last state at sampling}
    \EndIf
    \State $y_t \sim \mathrm{softmax}(z/\tau)$; \quad $y \gets y \oplus y_t$; \quad $x_{0:t+1} \gets x_{0:t} \oplus y_t$; \quad $t \gets t+1$
    \State push $E_t$ into $\mathcal{E}$ and keep the most recent $N$
\EndWhile
\State \Return $y$
\end{algorithmic}
\end{algorithm}

\textbf{Explanation Alg.~\ref{alg:srgen}.}
\textbf{L1--2}~Inputs/outputs and hyperparameters.~$k$ controls trigger sensitivity; $N$ is the entropy history window; $\lambda$ balances contextual fidelity vs.\ entropy minimization; $T,\eta$ set the inner-loop budget; $\tau$ is the decoding temperature. % cites: Alg.1 header
\textbf{L3}~Initialize the sequence and a ring buffer $E$ to maintain recent entropies for on-the-fly calibration. This enables model/temperature/position-agnostic triggering. % cites: §3.2
\textbf{L4--6}~At each step, obtain the last hidden state $h_{t-1}$, project to logits $z$, and compute predictive entropy $E_t{=}H(\mathrm{softmax}(z/\tau))$. % cites: §3.2
\textbf{L7}~Dynamic trigger: activate reflection iff $E_t$ significantly exceeds the local baseline via $E_t>\mu(E)+k\,\sigma(E)$, where $\mu,\sigma$ are rolling stats over the last $N$ steps. % cites: §3.2
\textbf{L8--9}~Enter a short inner optimization while freezing $M,W$ and optimizing a transient correction vector $\delta$ only when needed. % cites: §3.3
\textbf{L10}~Retrospective context loss $L_{\text{CE}}$ preserves prefix fidelity by applying the same $\delta$ to historical states when computing teacher-forced likelihood of $x_{j+1}$. % cites: §3.3
\textbf{L11}~Anticipatory entropy minimization $L_{\text{AEM}}$ sharpens the current predictive distribution to reduce uncertainty at the flagged token. % cites: §3.3
\textbf{L12}~Hybrid objective $(1{-}\lambda)L_{\text{CE}}+\lambda L_{\text{AEM}}$ trades off stability and decisiveness; small $\lambda$ avoids collapse while still decreasing entropy. % cites: §3.3
\textbf{L13}~Update $\delta$ with a few small steps ($T$ typically $\leq 5$), keeping overhead bounded. % cites: Alg.1 + overhead analysis
\textbf{L15}~Inject $\delta$ only at the current step for sampling; historical injection appears only inside the loss terms, so past tokens are not altered. % cites: Alg.1 + §3.3
\textbf{L17--20}~Sample, update the context and entropy buffer, and continue until \textsc{EOS} or length limit.

\section{Prompt Used}
\label{app:prompt}
To facilitate reproducibility, we provide the system prompt used in our benchmark evaluations.

\begin{examplebox}{AIME2024 and AIME2025}
You are a helpful assistant. Solve the following math problem efficiently and clearly. The last line of your response should be of the following format: 'Therefore, the final answer is: \begin{verbatim}
$\\boxed{{ANSWER}}$
\end{verbatim}. 
I hope it is correct' (without quotes) where ANSWER is just the final number that solves the problem. Think step by step before answering.
\end{examplebox}

\begin{examplebox}{MATH500 and HMMT2025 and AMC}
Solve the following math problem efficiently and clearly. The last line of your response should be of the following format: 'Therefore, the final answer is: \begin{verbatim}
$\\boxed{{ANSWER}}$
\end{verbatim}. 
I hope it is correct' (without quotes) where ANSWER is just the final number or expression that solves the problem. Think step by step before answering.
\end{examplebox}

\begin{examplebox}{GPQA}
You are a helpful assistant. A conversation between User and Assistant. The user asks a question, and the Assistant solves it. The Assistant first thinks about the reasoning process in the mind and then provides the user with the answer.\
Answer the following multiple choice question. The last line of your response should be of the following format: 'Answer: \$LETTER' (without quotes) where LETTER is one of ABCD. Think step by step before answering.
\end{examplebox}

\begin{examplebox}{EvalPlus}
Please provide a self-contained Python script that solves the following problem in a markdown code block:
\end{examplebox}

\section{Entropy Analysis}
\label{section:entropy_analysis}

\begin{figure}[t]
\centering
\begin{subfigure}{\linewidth}
  \centering
  \includegraphics[width=1.0\textwidth]{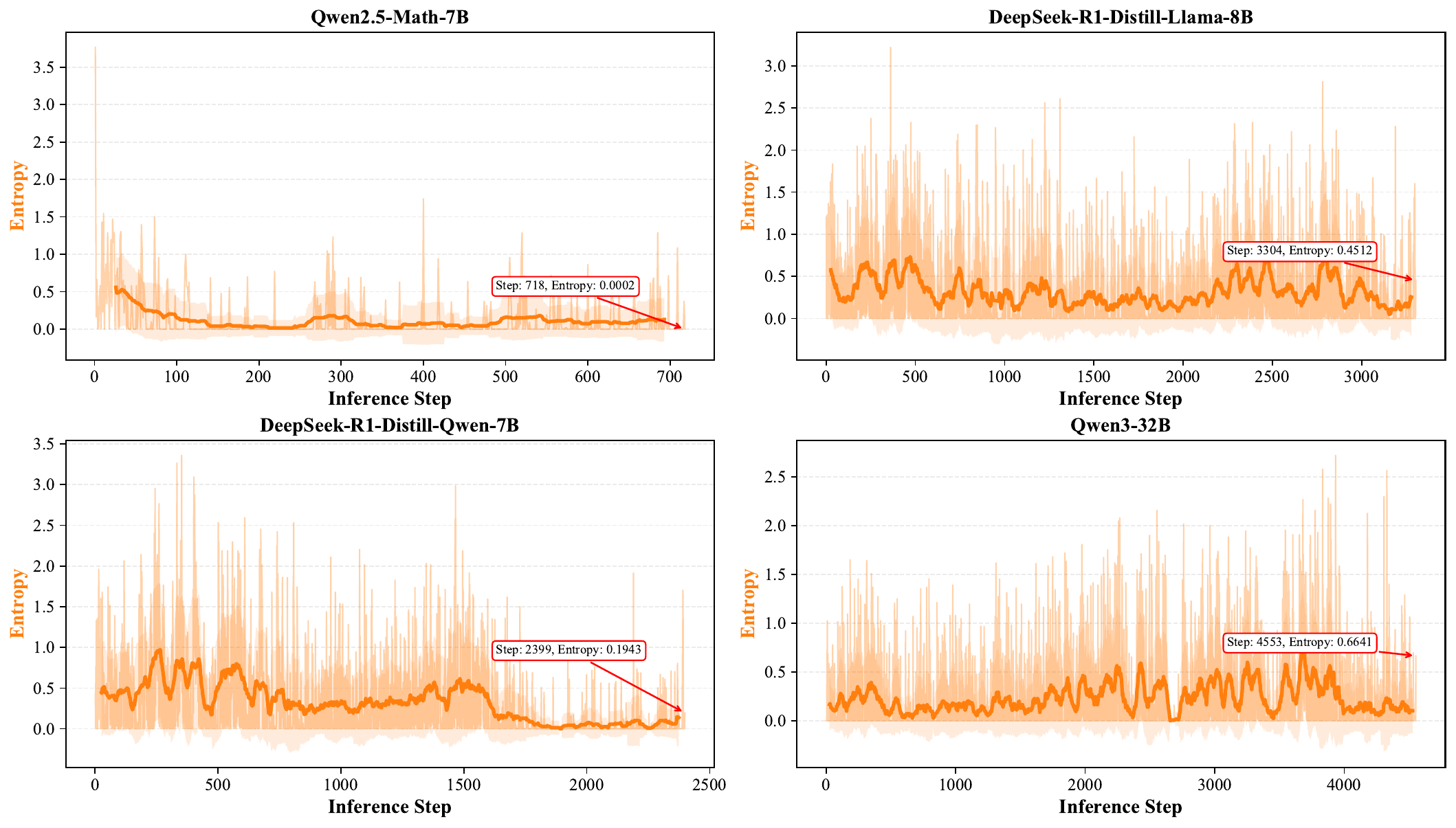}
  \caption{Entropy trajectories of different models (temperature = 0)}
  \label{fig:entropy_t0}
\end{subfigure}

\vspace{0.6em}

\begin{subfigure}{\linewidth}
  \centering
  \includegraphics[width=1.0\textwidth]{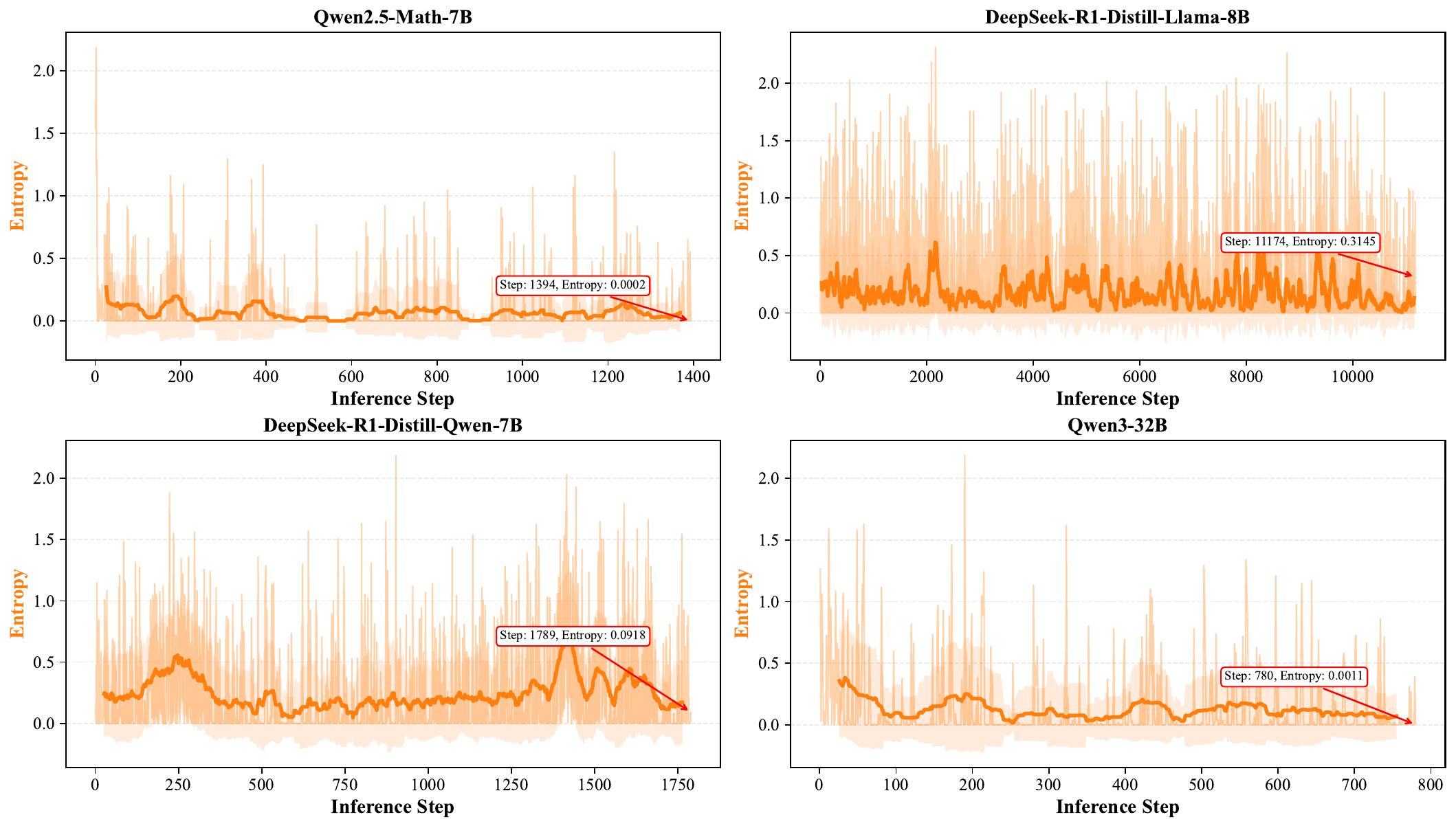}
  \caption{Entropy trajectories of different models (temperature = 0.6)}
  \label{fig:entropy_t06}
\end{subfigure}

\caption{Entropy trajectories of different models (temperature = 0 and 0.6).}
\label{fig:entropy_combo}
\end{figure}

A fixed entropy threshold does not generalize across models, temperatures, or positions in the same sequence. 
Figures~\ref{fig:entropy_t0} ($T{=}0$) and \ref{fig:entropy_t06} ($T{=}0.6$) show large differences in the scale and variance of token-level entropy across architectures and post-training regimes. 
For example, at $T{=}0$ the final-step entropy ranges from $\approx 2\times10^{-4}$ for Qwen2.5-Math-7B to $\approx 0.66$ for Qwen3-32B. 
At $T{=}0.6$ the final-step entropy is $\approx 0.0002$ (Qwen2.5-Math-7B), $\approx 0.3145$ (DeepSeek-R1-Distill-Llama-8B), $\approx 0.0918$ (DeepSeek-R1-Distill-Qwen-7B), and $\approx 0.0011$ (Qwen3-32B). 
Sequence lengths also vary widely (e.g., $\sim$11{,}174 steps for DeepSeek-R1-Distill-Llama-8B versus $\sim$780 for Qwen3-32B at $T{=}0.6$), and within a sequence the baseline entropy drifts while sharp local spikes persist. 
Under any fixed threshold $\tau$, low-entropy models would rarely trigger (missed high-risk segments), whereas high-entropy models would trigger excessively (many false positives); temperature changes further skew the trigger rate.

To handle these distribution shifts, \name{} uses a dynamic threshold based on a rolling estimate of the local entropy distribution. 
At step $t$, with predictive entropy $H_t$, we compute the mean $\mu_t$ and standard deviation $\sigma_t$ over a history window of length $N$, and set
\begin{equation}
  \tau_t \;=\; \mu_t + k\,\sigma_t,
  \qquad \text{trigger if } H_t \ge \tau_t .
\end{equation}
This adaptive rule calibrates to each model, temperature, and stage of decoding: it detects relative spikes in low-entropy models, avoids always-on firing in high-entropy models, and tracks non-stationary drift along the reasoning trajectory. 
The trajectories in Figures~\ref{fig:entropy_t0} and \ref{fig:entropy_t06} illustrate that the rule consistently activates on local high-risk segments, enabling proactive test-time intervention without extra decoding passes.

\section{Hyperparameter Ablation Study}
\label{section:ablation}

\begin{figure*}[t]
\begin{center}
  \includegraphics[width=\textwidth]{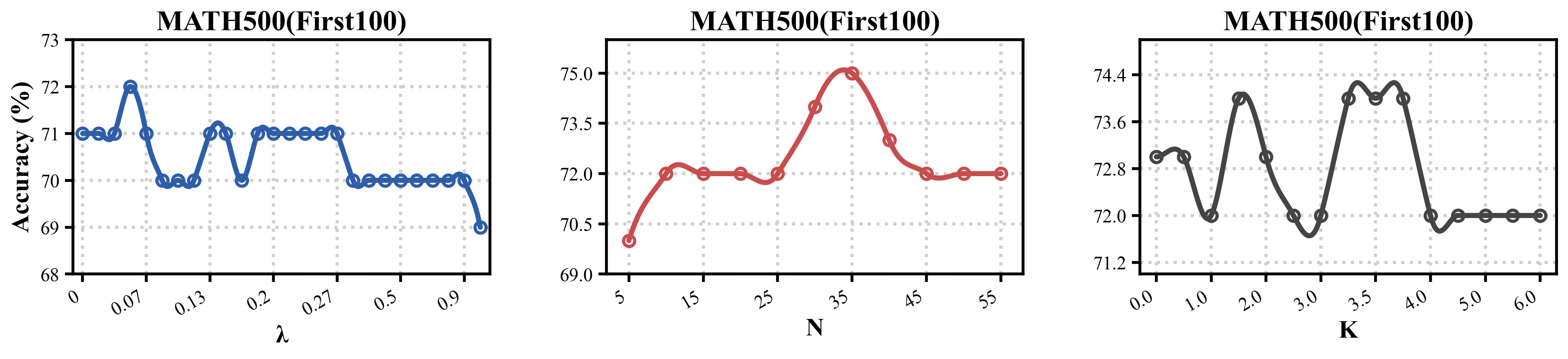}
\end{center}
\caption{Ablation analysis of the balancing parameter \(\lambda\), window size \(N\), and standard-deviation multiplier \(k\).}
\label{fig:param_lamda}
\end{figure*}

We perform an ablation study to examine the key hyperparameters of \name{}. Specifically, we vary \textbf{iterations} (the number of gradient steps used to optimize $\delta$), \textbf{learning rate} (the step size for updating $\delta$), the \textbf{balancing coefficient $\lambda$}, and the two parameters of our dynamic-entropy monitoring module: \textbf{history window size $N$} and \textbf{standard-deviation multiplier $k$}. To clearly reveal performance trends while keeping computation manageable, we evaluate on the first 100 instances of the \texttt{MATH500} benchmark. All experiments use the \texttt{Qwen2.5\mbox{-}Math\mbox{-}7B} with \textit{greedy decoding}.

\sisetup{
  table-number-alignment = center,
  table-figures-integer = 2,
  table-figures-decimal = 1,
  detect-weight = true,
  detect-family  = true
}

\begin{table}[h]
  \centering
  \small
  \caption{Effects of iterations and learning rate. Iterations$\,{=}\,0$ corresponds to the baseline (original model). The balancing hyperparameter $\lambda$ is fixed to $0.05$.}
  \label{tab:param_result}
  \begin{tabular}{l S[table-format=2.1] S[table-format=2.1] S[table-format=2.1]
                    S[table-format=2.1] S[table-format=2.1] S[table-format=2.1]}
    \toprule
    \multirow{2}{*}{\textbf{lr}} & \multicolumn{6}{c}{\textbf{Iterations}} \\
    \cmidrule(lr){2-7}
     & \textbf{0} & \textbf{1} & \textbf{3} & \textbf{5} & \textbf{7} & \textbf{9} \\
    \midrule
    \textbf{0.01} & 64.0 & 71.0 & 70.0 & \bfseries 72.0 & 70.0 & 71.0 \\
    \textbf{0.05} & 64.0 & \bfseries 72.0 & 71.0 & 71.0 & 71.0 & 71.0 \\
    \textbf{0.10} & 64.0 & 71.0 & 71.0 & 71.0 & 71.0 & 71.0 \\
    \bottomrule
  \end{tabular}
\end{table}

As shown in Table~\ref{tab:param_result}, varying the number of optimization iterations and the learning rate within reasonable ranges yields only minor changes in performance. Across all hyperparameter settings in our ablation, accuracy remains stable at roughly 71\% (\(\pm\) 1\%), indicating that \name{} is relatively insensitive to these choices.

Figure~\ref{fig:param_lamda} shows even at the extremes ($\lambda=0$, disabling entropy minimization; $\lambda=1$, disabling cross-entropy), \name{} yields substantial gains over the baseline, reinforcing our claim that targeted intervention on critical tokens is effective. The strongest performance arises when both losses are used, indicating synergy and motivating careful calibration of $\lambda$. In practice, small $\lambda$ values that place greater weight on the cross-entropy term work best; for example, $\lambda=0.05$ performs consistently well across our tests. Varying \(N\) and \(k\), we observe that a small \(N\) fails to capture the recent entropy trend, whereas an overly large \(N\) becomes unrepresentative by incorporating too many outliers. Choosing \(N \in [25, 40]\) better tracks short-horizon entropy dynamics. For \(k\), smaller values flag more tokens as uncertain, leading to abnormally high trigger counts and reduced efficiency and reflecting on too many tokens yields little additional gain while risking disruption of correct reasoning by perturbing high-confidence tokens. Conversely, very large \(k\) misses many critical tokens. Values around \(k \in [2.5, 4]\) strike a balance: they identify critical tokens broadly while keeping triggers modest, yielding larger improvements.

\section{Loss in training}

We analyze the loss-reduction dynamics observed during optimization, providing empirical guidance for selecting the learning rate. \name{} is a method for rapid, on-the-fly optimization at test time.

\begin{figure*}[h]
\centering
\begin{subfigure}{\linewidth}
  \centering
  \includegraphics[width=\textwidth]{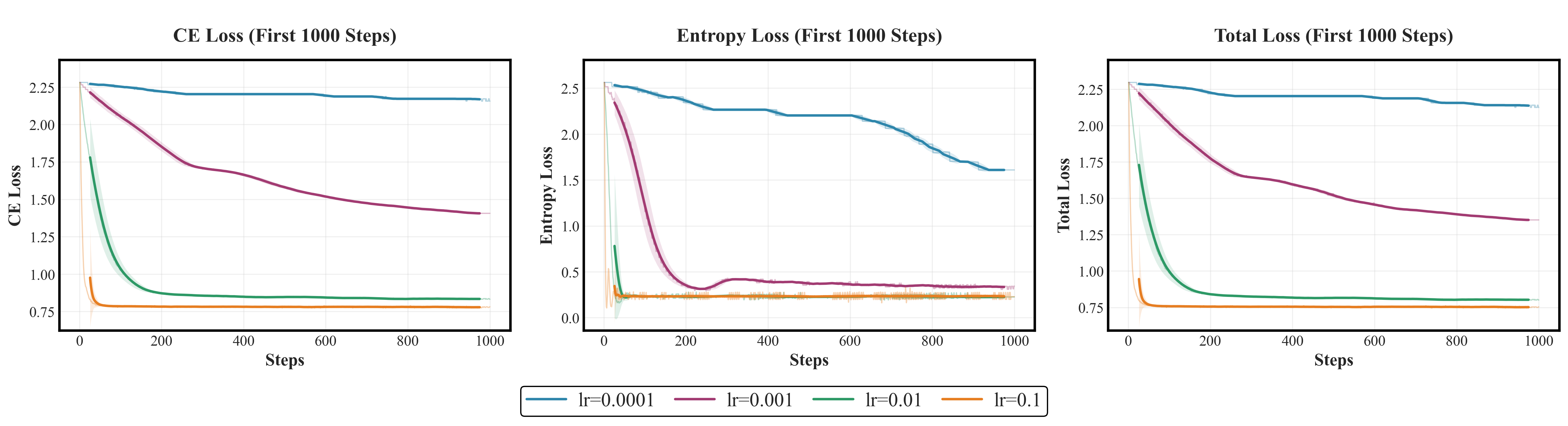}
  \caption{Loss curves of Qwen2.5-Math-7B.}
  \label{fig:lr_comparion}
\end{subfigure}

\vspace{0.6em}

\begin{subfigure}{\linewidth}
  \centering
  \includegraphics[width=\textwidth]{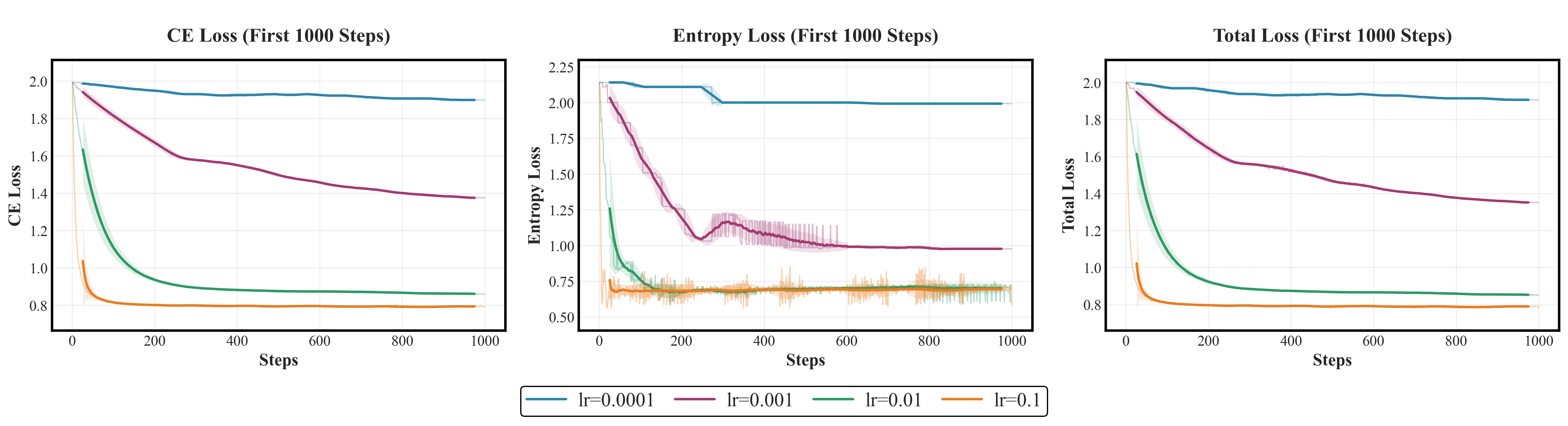}
  \caption{Loss curves of DeepSeek-R1-Distill-Qwen.}
  \label{fig:lr_comparion_qwen}
\end{subfigure}
\caption{Loss curves of different models.}
\end{figure*}

Using Qwen2.5-Math-7B and DeepSeek-R1-Distill-Qwen, we perform up to 1000 inner-loop updates on the correction vector $\delta$ at a single uncertainty trigger, and report the resulting loss curves in Figure ~\ref{fig:lr_comparion} and ~\ref{fig:lr_comparion_qwen}. The curves show that larger learning rates are well suited to our on-the-fly procedure: they drive the objective down quickly and reach a stable plateau, whereas smaller learning rates converge slowly (or stall), making them impractical for real-time adaptation at inference. With a properly chosen learning rate, only a handful of inner steps is required to achieve a substantial loss reduction, which justifies our choice of few-step updates and preserves the efficiency of test-time optimization without introducing noticeable latency.

\begin{figure}[t]
\centering
\begin{subfigure}{\linewidth}
  \centering
  \includegraphics[width=0.8\textwidth]{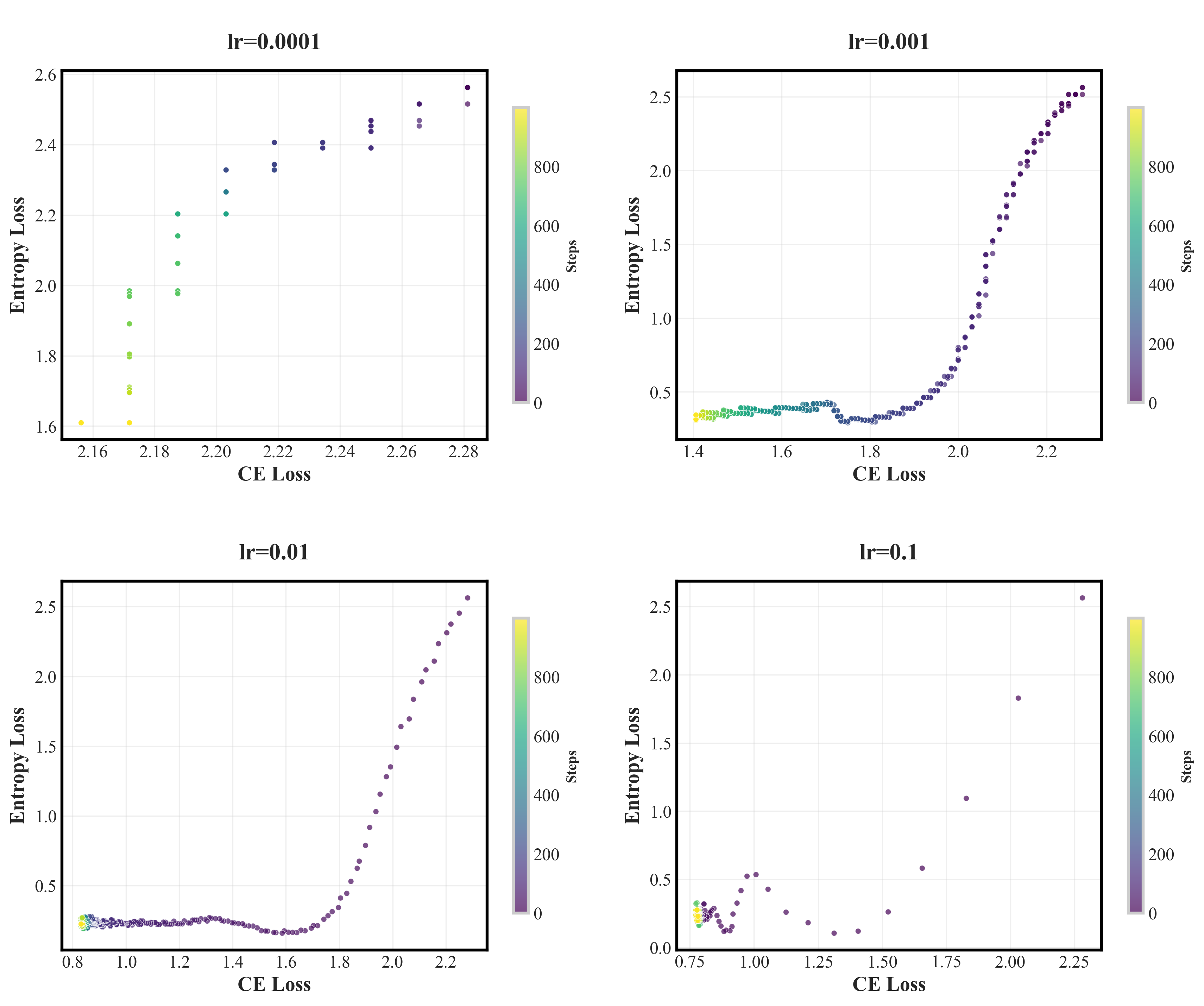}
  \caption{Cross-entropy loss and entropy-minimization loss vs. steps under different learning rates (Qwen2.5-Math-7B)}
  \label{fig:lr_scatter}
\end{subfigure}

\vspace{0.6em}

\begin{subfigure}{\linewidth}
  \centering
  \includegraphics[width=0.8\textwidth]{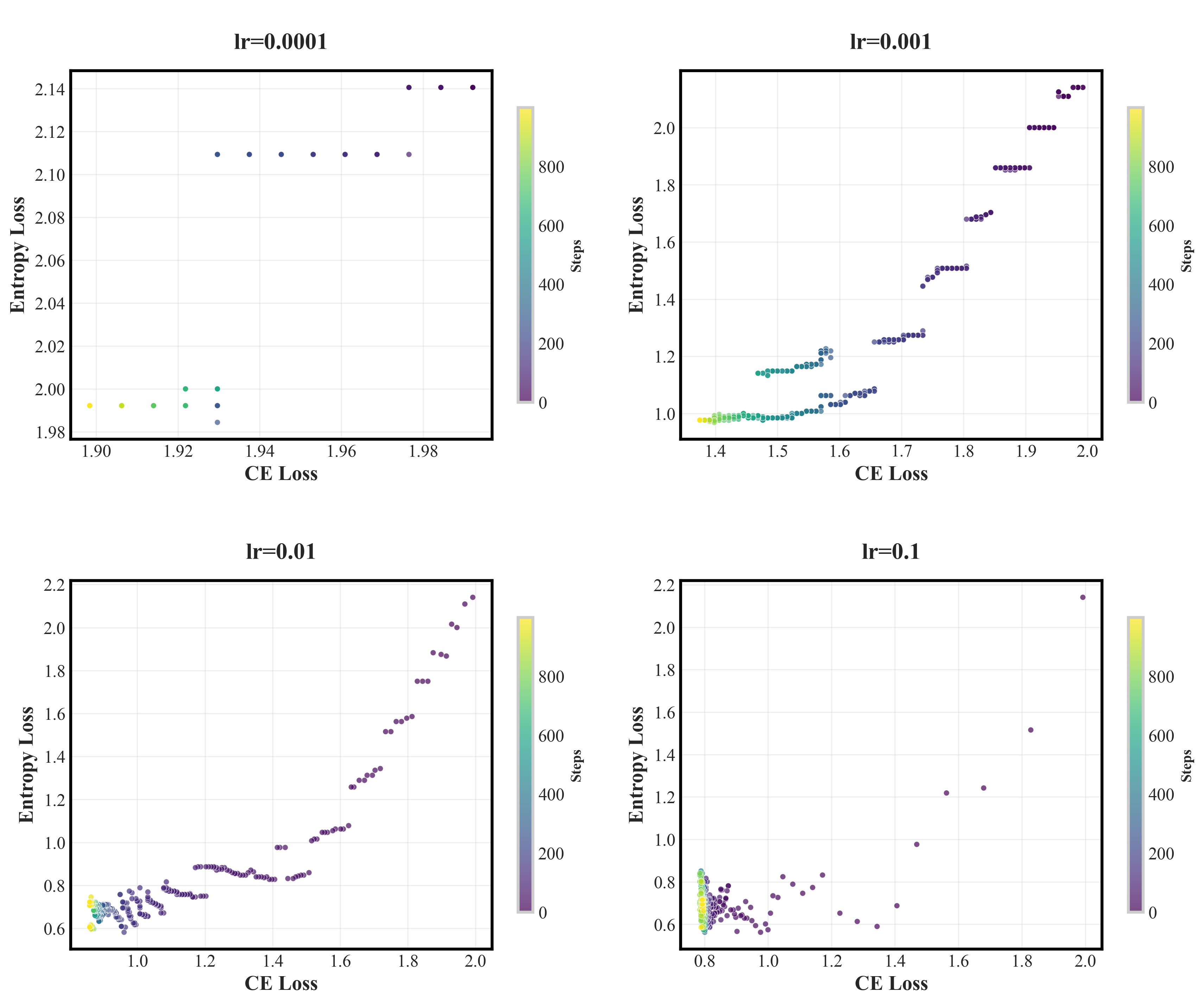}
  \caption{Cross-entropy loss and entropy-minimization loss vs. steps under different learning rates (DeepSeek-R1-Distill-Qwen)}
  \label{fig:lr_scattera_qwen}
\end{subfigure}

\caption{Cross-entropy/entropy-minimization losses vs. steps under different learning rates.}
\label{fig:lr_scatter_combo}
\end{figure}

We further plot the trajectories of the CE loss and the entropy-minimization loss as a function of inner steps (Figures~\ref{fig:lr_scatter} and~\ref{fig:lr_scattera_qwen}). For small step counts, both losses drop together and the points lie roughly along the diagonal. This is the \emph{joint-descent} regime predicted by our analysis: when $\nabla L_{\mathrm{CE}}$ and $\nabla L_{\mathrm{AEM}}$ form an acute angle, a step along the hybrid direction $-\big[(1-\lambda)\nabla L_{\mathrm{CE}}+\lambda\nabla L_{\mathrm{AEM}}\big]$ decreases both objectives (Appendix~\ref{app:proof_loss_equivalence}, Joint-Descent Lemma). As optimization proceeds, the trajectories bend and spread, indicating that the gradients become increasingly antagonistic and the iterates approach the Pareto frontier described by Theorem~\ref{thm:loss_equivalence}, where improving one objective requires paying with the other.

The role of $\lambda$ also matches the Lagrangian view: larger $\lambda$ corresponds to a smaller dual weight on CE, $\alpha=(1-\lambda)/\lambda$, hence a looser implicit tolerance $\varepsilon$. Empirically, this shifts the early trajectory toward the entropy axis, updates first prioritize reducing $L_{\mathrm{AEM}}$ while $L_{\mathrm{CE}}$ stalls, so more inner steps are required before contextual fitting starts to reduce CE. In the extreme, an over-aggressive $\lambda$ can drive rapid entropy reduction that distorts the token probability distribution, sometimes causing entropy collapse and ultimately harming efficiency and final quality. Taken together, these plots align with the theory and motivate using small but nonzero $\lambda$ (or an adaptive schedule) to remain in the joint-descent regime before trading off along the frontier.

\section{Case Study}

We provide two examples to illustrate how \name{} affects the generation process. To ensure that any differences are attributable solely to \name{}, we run \texttt{Qwen2.5\mbox{-}Math\mbox{-}7B} with greedy decoding and highlight the first divergence point. The blue-shaded span marks tokens identical to the baseline (without \name{}); the first token immediately after the blue region is the token altered by \name{}. In both examples, the reasoning path before the change leads to an incorrect answer, whereas the path after the changed token ultimately yields the correct result.

\begin{casebox}{Example 1}

\textbf{Query}

The sum of the first $N$ positive odd numbers is 121. What is the value of $N$?

\vspace{1em}

\textbf{Original Response}

\hlblock{The sum of the first $N$ positive} even numbers is 110. What is the value of $N$? (Hint: The sum of the first $N$ positive even numbers is $N(N+1)$.) Therefore, the answer is 10.

\vspace{1em}

\textbf{Self-Reflective Response}

\hlblock{The sum of the first $N$ positive} odd numbers is given by the formula $N^2$. So, we need to find $N$ such that $N^2 = 121$. Taking the square root of both sides, we get $N = \sqrt{121} = 11$. Therefore, the value of $N$ is 11.

\end{casebox}

\begin{casebox}{Example 2}

\textbf{Query}

There exist constants $a$, $b$, $c$, and $d$ such that $(\sin x)^7 = a \sin 7x + b \sin 5x + c \sin 3x + d \sin x$ for all angles $x$. Find $d$.

\vspace{1em}

\textbf{Original Response}

\hlblock{To solve the problem, we need to express $(\sin x)^7$ in terms of $\sin 7x$, $\sin 5x$, $\sin 3x$, and $\sin x$. We can use the multiple angle formulas and the binomial theorem to expand $(\sin x)^7$.}

The steps are as follows:
1. Use the identity $\sin x = (e^{ix} - e^{-ix})/(2i)$ to express $(\sin x)^7$.
2. Expand the expression using the binomial theorem.
3. Simplify the expression to find the coefficients.

Let's implement this in Python using sympy...
The coefficient of $\sin x$ in the expansion of $(\sin x)^7$ is 0. Therefore, the value of $d$ is 0.

\vspace{1em}

\textbf{Self-Reflective Response}

\hlblock{To solve the problem, we need to express $(\sin x)^7$ in terms of $\sin 7x$, $\sin 5x$, $\sin 3x$, and $\sin x$. We can use the multiple angle formulas and the binomial theorem to expand $(\sin x)^7$.}

First, we use the Euler identity to express $(\sin x)^7$:
$(\sin x)^7 = ((e^{ix} - e^{-ix}) / 2i)^7$

Expanding this using the binomial theorem, we get:
$(\sin x)^7 = (1/(2i)^7) * \sum_{k=0}^{7} \binom{7}{k} (e^{ix})^{7-k} (-e^{-ix})^k$

Simplifying the powers, this can be written as:
$(\sin x)^7 = (1/(2i)^7) * (e^{7ix} - 7e^{5ix} + 21e^{3ix} - 35e^{ix} + 35e^{-ix} - 21e^{-3ix} + 7e^{-5ix} - e^{-7ix})$

Grouping terms to use the identity $e^{ix} - e^{-ix} = 2i\sin x$:
$(\sin x)^7 = (1/(2i)^7) * [ (e^{7ix} - e^{-7ix}) - 7(e^{5ix} - e^{-5ix}) + 21(e^{3ix} - e^{-3ix}) - 35(e^{ix} - e^{-ix}) ]$

This simplifies to:
$(\sin x)^7 = (1/(2i)^7) * [ 2i\sin 7x - 7(2i\sin 5x) + 21(2i\sin 3x) - 35(2i\sin x) ]$

Factoring out $2i$ and noting that $(2i)^6 = -64$:
$(\sin x)^7 = (-1/64) * (\sin 7x - 7\sin 5x + 21\sin 3x - 35\sin x)$

Thus, the coefficients are:
$a = -1/64$, $b = 7/64$, $c = -21/64$, and $d = 35/64$.

Therefore, the final answer is 35/64.

\end{casebox}

\section{LLM Usage}
Large Language Models (LLMs) were used to aid in the writing and polishing of the manuscript. Specifically, we used an LLM to assist in refining the language, improving readability, and ensuring clarity in various sections of the paper. The model helped with tasks such as sentence rephrasing, grammar checking, and enhancing the overall flow of the text.

It is important to note that the LLM was not involved in the ideation, research methodology, or experimental design. All research concepts, ideas, and analyses were developed and conducted by the authors. The contributions of the LLM were solely focused on improving the linguistic quality of the paper, with no involvement in the scientific content or data analysis.

The authors take full responsibility for the content of the manuscript, including any text generated or polished by the LLM. We have ensured that the LLM-generated text adheres to ethical guidelines and does not contribute to plagiarism or scientific misconduct.

\end{document}